\documentclass[sn-apa]{sn-jnl}

\usepackage{amsfonts} 
\usepackage{amsmath} 
\usepackage{amssymb} 
\usepackage[title]{appendix} 
\usepackage[utf8]{inputenc} 
\usepackage[T1]{fontenc} 
\usepackage[style = apa, backend = biber]{biblatex} 
\usepackage{booktabs} 
\usepackage{glossaries-extra} 
\usepackage{graphicx} 
\usepackage{latexcolors} 
\usepackage{lmodern} 
\usepackage{microtype} 
\usepackage{multirow} 
\let\orcidlogo\relax
\usepackage{orcidlink} 
\usepackage{pgfplots} 
\usepackage{subcaption} 
\usepackage{tabu} 
\usepackage{xcolor} 
\usepackage{tikz} 
\usetikzlibrary{arrows, arrows.meta, calc, decorations.pathreplacing, matrix, positioning, shapes} 

\newcommand{\I}{\mathcal{I}}

\newcommand{\buckets}{\mathcal{B}}

\newcommand{\bucket}[1]{\mathcal{B}_{#1}}

\newcommand{\B}{\succeq_{\mathcal{B}}}

\renewcommand{\vec}[1]{\boldsymbol{#1}}

\newcommand{\given}{\, | \,}

\newcommand{\haty}{\hat{y}}

\newcommand{\N}{\mathbb{N}}

\newcommand{\R}{\mathbb{R}}

\newcommand{\id}{1\hspace{-0,9ex}1}

\newcommand{\round}{\mathrm{round}}

\newcommand{\norm}{\mathrm{norm}}

\newcommand{\cX}{\mathcal{X}}

\newcommand{\cM}{\mathcal{M}}
\newcommand{\cD}{\mathcal{D}}

\newcommand{\x}{\vec{x}} 

\newcommand{\na}{\mathrm{NA}}

\newcommand{\Prob}{\mathbb{P}}

\newcommand{\argmin}{\operatorname*{argmin}}

\newcommand{\bO}{\mathcal{O}}

\setabbreviationstyle[acronym]{long-short}
\glsdisablehyper 

\newacronym{eps}{EPS}{\(\epsilon\)-closeness}
\newacronym{llm}{LLM}{large language model}
\newacronym{lr}{LR}{label ranking}
\newacronym{mor}{MOR}{multi-output regression}
\newacronym{more-plr}{MORE-PLR}{multi-output regression employed for partial label ranking}
\newacronym{obop}{OBOP}{optimal bucket order problem}
\newacronym{pi}{PI}{prediction interval}
\newacronym{plr}{PLR}{partial label ranking}
\newacronym{rpc}{RPC}{ranking by pairwise comparison}
\newacronym{rr}{RR}{round-rank}
\newacronym{st}{ST}{single-target}

\usepackage{style}

\begin{document}

\title{An extension of ``MORE--PLR: Multi-Output Regression Employed for Partial Label Ranking''}

\author[1]
{%
    \orcidlink{0009-0006-4163-6699} %
    \fnm{Santo M. A. R.}%
    \sur{Thies}%
}%
\email{S.Thies@campus.lmu.de}%

\author*[3,4]
{%
    \orcidlink{0000-0003-1777-8540} %
    \fnm{Juan C.}%
    \sur{Alfaro}%
}%
\email{JuanCarlos.Alfaro@uclm.es}

\author[1,2]
{%
    \orcidlink{0000-0001-6988-6186} %
    \fnm{Viktor}%
    \sur{Bengs}%
}%
\email{Viktor.Bengs@ifi.lmu.de}

\affil[1]
{
    \orgdiv{Chair of Artificial Intelligence and Machine Learning}, %
    \orgname{Ludwig-Maximilians-Universität München}, %
    \orgaddress
    {%
        \city{Munich}, %
        \postcode{80799}, %
        \country{Germany}%
    }%
}

\affil[2]
{
    \orgdiv{Munich Center for Machine Learning}, %
    \orgaddress
    {%
        \city{Munich}, %
        \country{Germany}%
    }%
}

\affil[3]
{
    \orgdiv{Departamento de Sistemas Informáticos}, %
    \orgname{Universidad de Castilla-La Mancha}, %
    \orgaddress
    {%
        \city{Albacete}, %
        \postcode{02071}, %
        \country{Spain}%
    }%
}

\affil[4]
{
    \orgdiv{Laboratorio de Sistemas Inteligentes y Minería de Datos}, %
    \orgname{Universidad de Castilla-La Mancha}, %
    \orgaddress
    {%
        \city{Albacete}, %
        \postcode{02071}, %
        \country{Spain}%
    }%
}

    \abstract
{
    The \gls{plr} problem is a supervised learning scenario that aims to fit a \textit{preference model} that predicts a \textit{bucket order} defined over a set of labels for a given input instance. 
    This problem generalizes the well-known \gls{lr} problem, which, in practice, is limited to outputting \textit{total orders} of labels.
    Existing \gls{plr} methods have primarily extended \gls{lr} approaches to handle \textit{ties} in predictions.
    This paper proposes using \gls{mor} to address the \gls{plr} problem, introducing an \textit{encoder} that, during the learning phase, transforms the (possibly incomplete) \textit{rankings with ties} of labels to multivariate regression targets, an underexplored perspective in both \gls{lr} and \gls{plr}.
    Moreover, during the inference phase, we introduce several \textit{post-hoc layers} that convert the \gls{mor} results into the output bucket order to effectively implement this approach.
    This framework provides learning strategies that are competitive with the current state-of-the-art \gls{plr}~methods, as demonstrated through experimental evaluations.
}


    \maketitle

    \section{Introduction}
    \label{section:introduction}
\begin{figure}[t]
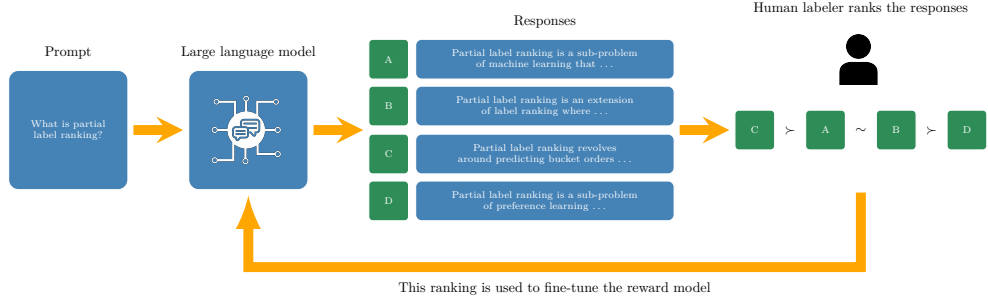

    \centering
    \include{figures/llm_tuning_example}
    \caption{Given a prompt, the \gls{llm} generates multiple responses, which are ranked by a human labeler from worst to best. This ranking is then used to fine-tune the \gls{llm}.}
    \label{figure:LLM-tuning}
\end{figure}

    In many machine learning applications, especially those involving human activities such as assessing medical treatments, conducting opinion polls, evaluating sports competitions, or developing recommender systems, the available data often consists of partial or entirely qualitative information rather than quantitative.
For example, consider the task of a human labeler during the fine-tuning step of a \gls{llm}: given a prompt, the \gls{llm} generates several responses, which the human labeler ranks from worst to best without assigning numerical scores to the outputs (see Fig. \ref{figure:LLM-tuning}).
This purely qualitative signal is then used to learn a suitable reward function for fine-tuning the \gls{llm}, ultimately improving alignment \parencite{kaufmann_survey_2024}. 

This scenario illustrates the \glsxtrfull{lr} problem \parencite{vembu_label_2010}, where the task is to learn a \textit{preference model} that predicts a \textit{total order} of available labels (responses) for a given input instance (prompt), aligning with an unknown ground-truth.
This specific learning task falls within the broader field of \textit{preference learning} \parencite{hullermeier_preference_2024}, with applications in \textit{multi-label classification} \parencite{furnkranz_multilabel_2008}, \textit{algorithm selection} \parencite{hanselle_hybrid_2020}, and \textit{monocular depth estimation} \parencite{lienen_monocular_2021}, among others.

Despite the widespread use of these learning methods, a practical limitation is that the model must output a total order of the labels.
As a result, existing approaches exclude the possibility of \textit{ties}, where multiple labels share the same ranking position.
For instance, in the \gls{llm} fine-tuning example mentioned earlier, a human labeler may assign the same position to two or more \gls{llm} outputs (see Fig. \ref{figure:LLM-tuning}).
This situation typically occurs when the outputs are considered equally \textit{good}, \textit{mediocre}, or \textit{bad} or when the human labeler lacks enough information to distinguish between them.

To address this limitation, the \gls{lr} problem was recently generalized to the \glsxtrfull{plr} problem \parencite{alfaro_learning_2021}, which allows training datasets to include (possibly incomplete) \textit{rankings with ties} (a.k.a. \textit{partial rankings}).
Additionally, in the PLR problem, the model can produce \textit{weak orders} (a.k.a. \textit{bucket orders}) instead of total orders.
So far, existing \gls{plr} methods have mostly extended \gls{lr} approaches, adapting them to accommodate ties in the output.
For instance, the decision tree approach proposed by \textcite{cheng_decision_2009} for the \gls{lr} problem was later extended to handle \gls{plr} datasets in the initial work by \textcite{alfaro_learning_2021}.

This paper introduces a meta-learning approach, previously explored only in a specialized form for \gls{lr} problems \parencite{cheng_labelwise_2013, fotakis_label_2022}.
The key idea is to treat the \textit{rank positions} as \textit{multivariate targets}, enabling any \glsxtrfull{mor} learner to serve as a \gls{plr} learner.
To facilitate this, we first apply an \textit{encoding} step that transforms the bucket orders into a structured multivariate representation, ensuring compatibility with \gls{mor} learners.
However, during the inference phase, the predicted multivariate vector may not directly correspond to a rank position vector, preventing direct reconstruction of the encoded bucket order.
Therefore, an additional transformation step or \gls{plr} post-hoc layer is needed.
To this end, we propose several \gls{plr} post-hoc layers, each designed to convert an arbitrary multivariate vector into a bucket order, thereby providing the prediction for a given input instance.
This approach defines a new class of methods, termed \gls{more-plr}, where each model consists of three key components: an encoding module, a base \gls{mor} learner, and a \gls{plr} post-hoc layer.

Compared to the state-of-the-art \textit{pairwise comparison reduction} method \parencite{alfaro_pairwise_2023}, which has a quadratic runtime concerning the number of labels, the \gls{more-plr} approach typically exhibits a linear runtime dependency.
Our experimental results show that the \gls{more-plr} framework achieves satisfactory accuracy on benchmark datasets, depending on the choice of encoder, \gls{mor} learner and \gls{plr} post-hoc layer.

The paper is organized as follows. Section \ref{section:related_work} provides an overview of the related work.
Section \ref{section:preliminaries} formally defines the \gls{plr} and the \gls{mor} problems. 
Section \ref{section:more-plr} describes the class of \gls{more-plr} learners, focusing on the different encoders and \gls{plr} post-hoc layers.
Experiments using benchmark datasets and state-of-the-art \gls{plr} methods are presented in Section \ref{section:experiments}, followed by our main conclusions in Section \ref{section:conclusions}. 

\paragraph{Extension of the conference version}

Our previous paper introduced the novel class of methods we termed \gls{more-plr}.
In the initial design of the framework, each model consisted of only two components: a base \gls{mor} learner and a \gls{plr} post-hoc layer, with a fixed approach for transforming the \gls{plr} dataset into a \gls{mor} one.
This paper introduces the encoder module that enables different methods for this transformation.
In addition to the original encoder, we propose three additional encoder variants and a new post-hoc layer, the \(\epsilon\)-closeness layer.
With three components now featuring different hyperparameter configurations, the experimental evaluation presented here is far more extensive than in the first version of the framework.
This allows us to identify several promising combinations of these components for specific learning scenarios in this domain.

    \section{Related Work}
    \label{section:related_work}
    This section reviews the most significant works on \gls{lr} and \gls{plr} learning scenarios.

\subsection*{Label Ranking}

Three main types of approaches are commonly employed to address the \gls{lr} problem \parencite{zhou_taxonomy_2014}.
The first category applies a \textit{reduction} technique that transforms the \gls{lr} problem into a series of \textit{classification} tasks, which are then combined to make the prediction \parencite{cheng_labelwise_2013, har-peled_constraint_2002, hullermeier_label_2008, vogel_multiclass_2020}.
The second variant adapts classic machine learning algorithms, such as \textit{decision trees} or \textit{neural networks}, to directly model the \gls{lr} structure \parencite{cheng_decision_2009, ribeiro_multilayer_2012, de_sa_mining_2011, zhou_label_2014}.
The third kind employs \textit{ensemble} methods, where multiple models are trained, and their predictions are aggregated for each input instance \parencite{aledo_tackling_2017, dery_boostlr_2020, de_sa_label_2017}.
The idea of using \gls{mor}, as proposed in this paper, shares similarities with the work of \textcite{fotakis_label_2022}.
Specifically, a \textit{single-target} approach was applied with \textit{tree-based} learners, where outputs were subsequently sorted to obtain the predicted total order. This method represents a particular case of our proposed \gls{more-plr} framework.
However, the effectiveness of this rather simplistic approach remains experimentally uncertain, as it fails to capture dependencies between ranking positions (see Section \ref{subsection:mor}).
Instead, the authors focus on theoretical aspects, such as sample complexity bounds and Bayes errors.
Another similar idea was explored by \textcite{cheng_labelwise_2013}, though their work focused on restricted ordinal classification rather than \gls{mor}.

\subsection*{Partial Label Ranking}

Methods for the \gls{plr} problem typically build upon learning algorithms originally designed for \gls{lr}, extending them to accommodate ties in both training data and predictions.
Several methods have been adapted, including decision trees \parencite{cheng_decision_2009}, ensemble learning approaches \parencite{aledo_tackling_2017}, and the pairwise comparison reduction technique \parencite{har-peled_constraint_2002}, with further refinements for \gls{plr} \parencite{alfaro_pairwise_2023, alfaro_ensemble_2022}.
An instance-based \gls{plr} approach also follows the classical \textit{nearest neighbors} classification principle \parencite{alfaro_learning_2021}.

    \section{Preliminaries}
    \label{section:preliminaries}
    This section covers the key theoretical concepts needed to understand the paper.
Section \ref{subsection:plr} provides a formal introduction to the \glsxtrfull{plr} problem, explaining the relevant concepts and terminology.
Similarly, Section \ref{subsection:mor} presents the \glsxtrfull{mor} framework.

        \subsection{Partial Label Ranking}
        \label{subsection:plr}
        Let \(\I = \{u_1, \ldots, u_k\}\) denote a set of items, where \(k \in \mathbb{N}\).
An ordered partition of \(\I\) into non-empty disjoint subsets, denoted as \(\buckets = \langle\bucket{1}, \ldots, \bucket{c}\rangle\), with \(c \in [k] = \{1, \ldots, k\}\), \(\bigcup_{l = 1}^{c} \bucket{l} = \I\), and \(\bucket{l} \subseteq \I\), is referred to as a \textit{bucket order}.
It specifies a \textit{total order with ties} \(\B\) on \(\I\) as follows: if \(u_i \in \bucket{l_i}\) and \(u_j \in \bucket{l_j}\), then \(u_i \succ_{\buckets} u_j\) if and only if \(l_i < l_j\), while \(u_i \sim_{\buckets} u_j\) if and only if \(l_i = l_j\).
In other words, items belonging to the same \textit{bucket} are tied, while items in preceding buckets are ranked higher than (or preferred to) all items in subsequent buckets. 
Note that a total order is recovered when \(c = k\), meaning each bucket consists of exactly one item.

Another equivalent way of representing a bucket order \(\) is through its associated bucket matrix \(B =(B_{i, j})_{i, j=1}^k \in \{0, 0.5, 1\}^{k \times k}\), where each entry is computed as:
\begin{align}
    \label{def:bucket_matrix}
    B_{i,j} =  \begin{cases}
    1 & \text{if } u_i \succ_{\buckets} u_j, \\
    0.5 & \text{if } u_i \sim_{\buckets} u_j, \\
    0 & \text{otherwise.}
    \end{cases}
\end{align}
In other words, pairs of items belonging to the same bucket are assigned a \(0.5\) entry. Otherwise, a \(0\) or \(1\) value is assigned based on whether an item is in a subsequent or preceding bucket, respectively.

Given an \textit{instance} or \textit{feature space} \(\cX\), and treating the items in \(\I\) as \textit{labels}, the \glsxtrfull{lr} problem \parencite{zhou_taxonomy_2014} involves learning a mapping \(\cX \to \mathcal{S}_k^{\succ}\), where \(\mathcal{S}_k^{\succ}\) denotes the set of all total orders of the \(k\) labels in \(\I\).
This mapping, known as a \textit{label ranker}, is learned from a dataset \(\cD \subseteq \cX \times \mathcal{S}_k^{\succ,\mathrm{inc}}\). 
Here, \(\mathcal{S}_k^{\succ, \mathrm{inc}}\) is the set of (possibly incomplete) \textit{rankings} of \(\I\), considering that some labels may not be ranked for specific instances.
The goal of the label ranker is to predict a total order \(\succ_{\x}\) of \(\I\) for a given input instance \(\x \in \cX\).
This prediction establishes an ordering relationship between any two items in the form \(u_i \succ_{\x} u_j\), interpreted as \(u_i\) being preferred over \(u_j\) given the context \(\x\). 

When predicting a total order of items is infeasible or undesirable, the \glsxtrfull{plr} problem allows for bucket orders.
Formally, this entails a mapping \(\cX \to \mathcal{S}_k^{\succeq}\), called \textit{partial label ranker}, where \(\mathcal{S}_k^{\succeq}\) represents the set of all bucket orders of the of the \(k\) labels in \(\I\).
Learning is conducted based on a dataset \(\cD \subseteq \cX \times \mathcal{S}_k^{\succeq, \mathrm{inc}}\), where \(\mathcal{S}_k^{\succeq, \mathrm{inc}}\) denotes the set of (possibly incomplete) \textit{rankings with ties} (a.k.a. \textit{partial rankings}) of \(\I\).
Thus, unlike a typical \gls{lr} dataset, a \gls{plr} dataset may include rankings with tied labels for specific instances.

The most common approach to addressing the \gls{plr} problem is to train a \textit{classifier}, \(\cM\), to estimate the probabilities of each relationship between item pairs \((u_i, u_j) \in \I \times \I\) with \(i < j\).
These relationships include \(u_i \succ u_j\), \(u_i \sim u_j\), and \(u_i \prec u_j\), representing the relative position of \(u_i\) concerning \(u_j\).
The estimated probabilities are then used to construct a so-called \textit{pair order matrix} \(C \in [0,1]^{k \times k}\) as follows:
\begin{align*}
    C_{i, j} = \Prob(u_i \succ u_j \given \cM) + 0.5 \cdot \Prob(u_i \sim u_j \given \cM).
\end{align*}
Here, \(C_{i, j}\) represents the predicted probability of \(u_i\) being preferred over \(u_j\), plus half the predicted probability of \(u_i\) being tied with \(u_j\).
Since \(C\) does not necessarily correspond to a bucket matrix, it must be transformed into one to obtain the final prediction. 
This transformation is performed by solving the \gls{obop}, which aims to find the closest bucket matrix in terms of \(L_1\) distance to \(C\), that is:
\begin{equation*}
    \label{equation:obop}
    \argmin\nolimits_{B} \sum\nolimits_{u_i, u_j \in \I} \left|B_{i, j} - C_{i, j}\right|.
\end{equation*}
Several algorithms have been proposed to address the \gls{obop}; see \textcite{aledo_utopia_2017, ukkonen_randomized_2009} for an overview.

Figure \ref{figure:plr} illustrates the learning and prediction process in the \gls{plr} problem. 
As shown, whenever a prediction for an instance \(\x\) is requested, the classifier \(\cM\) is queried, followed by solving the \gls{obop} to obtain the pair order matrix \(C\).

        \subsection{Multi-Output Regression}
        \label{subsection:mor}
        Classical \textit{single-output regression} involves predictions of the form \(h: \cX \to \R\), meaning that for a given instance \(\x \in \cX\), the \textit{regressor} predicts a real-valued target \(\haty = h(\x)\).
For example, predicting the price of a car based on features like \(75\) horsepower, \(1462\) engine displacement, \(4\) cylinders, among others.
\Glsxtrfull{mor} is the \textit{multivariate} extension of classical single-output regression \parencite{borchani_survey_2015}, where predictions are of the form \(H:\cX \to \R^q\).
In other words, for a given instance \(\x \in \cX\), a real-valued multivariate target \(\vec{\haty} = H(\x)\) is predicted, with \(q \in \mathbb{N}\).
For instance, predicting a car's price and fuel consumption based on its features.

Based on the latter example, we can infer that in \gls{mor}, there may be relationships within the feature space, between the feature space and the target space, and within the target space itself.
In the previous example, there is undoubtedly some correlation between the price and a car's fuel consumption. 
Accordingly, most approaches for \gls{mor} extend single-output regression models to account for this potential additional correlation.
These methods can be classified into two categories.
First, \textit{tranformation} methods tackle the problem of constructing \gls{mor} models by combining multiple (standard) single-output models. 
This class of methods includes:
\begin{itemize}
    \item the \textit{single-target} method, which predicts each target variable independently, disregarding potential interdependencies;
    \item the \textit{stacking} method, which performs regression for each target variable initially, then uses these predictions as additional features in a subsequent step;
    \item the \textit{regression chain} method, which starts by defining an order (random or heuristic) for the target variables and then learns single-output regressors, incorporating the predictions of the previous learner(s) in the chain into the feature space for the prediction of the next target variable;
    \item and the \textit{multi-output support vector regression}, which transforms the feature space to handle \gls{mor} as a single-output regression problem.
\end{itemize}

The second class of approaches is \textit{algorithm adaptation} methods, which modify classical models to inherently support multi-output problems by capturing all relationships and dependencies among the outputs.
\textit{Multi-target Gaussian processes} \parencite{jakkala_deep_2021} and \textit{multi-target regression trees} \parencite{de_multivariate_2002} are two such algorithms proposed in the literature \parencite{borchani_survey_2015}, among others.

    \section{Multi-Output Regression Employed for Partial Label Ranking}
    \label{section:more-plr}
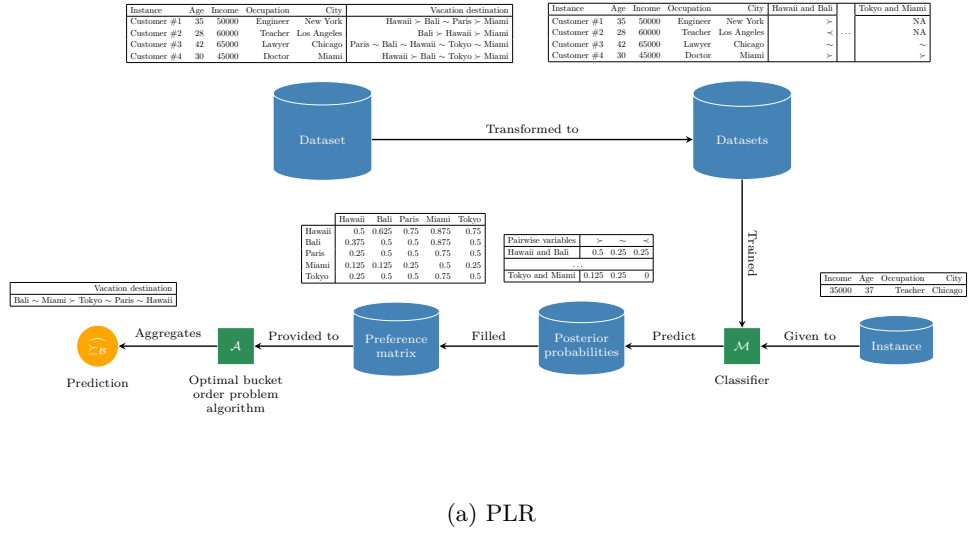
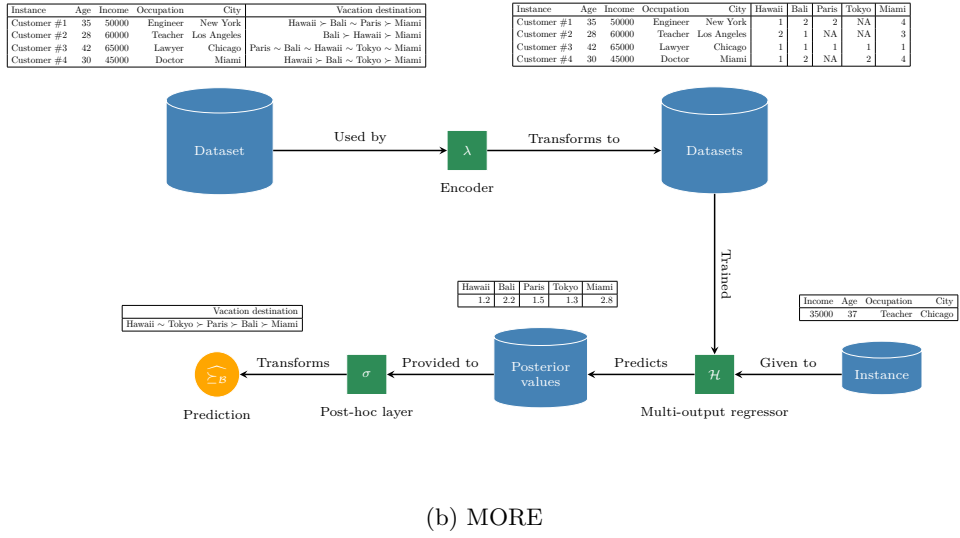
\begin{figure}[!ht]
    \centering
    \begin{subfigure}{\textwidth}
        \resizebox{\linewidth}{!}
{
    \begin{tikzpicture}[thick]

        \setlength{\tabcolsep}{0.15cm}
    
        \node (dataset) [dataset] {Dataset};
        \node [above = 0.25 of dataset]
        {
            \scalebox{0.5}
            {
                \centering
                \begin{tabu}{|lrrrr|r|}
                    \hline
                    Instance & Age & Income & Occupation & City & Vacation destination \\
                    \hline
                    Customer \#1 & 35 & 50000 & Engineer & New York & Hawaii \(\succ\) Bali \(\sim\) Paris \(\succ\) Miami \\
                    Customer \#2 & 28 & 60000 & Teacher & Los Angeles & Bali \(\succ\) Hawaii \(\succ\) Miami \\
                    Customer \#3 & 42 & 65000 & Lawyer & Chicago & Paris \(\sim\) Bali \(\sim\) Hawaii \(\sim\) Tokyo \(\sim\) Miami \\
                    Customer \#4 & 30 & 45000 & Doctor & Miami & Hawaii \(\succ\) Bali \(\sim\) Tokyo \(\succ\) Miami \\
                    \hline
                \end{tabu}
            }
        };
    
        \node (datasets) [dataset, right = 6.5cm of dataset] {Datasets};
        \path (dataset) edge node [annotation, above] {Transformed to} (datasets);
        \node [above = 0.25 of datasets]
        {
            \scalebox{0.5}
            {
                \centering
                \centering
                \begin{tabu}{|lrrrr|r|r|r|}
                    \hline
                    Instance & Age & Income & Occupation & City & Hawaii and Bali & \multirow{5}{*}{\(\dots\)} & Tokyo and Miami \\
                    \tabucline{1-6}
                    \tabucline{8-8}
                    Customer \#1 & 35 & 50000 & Engineer & New York & \(\succ\) & & \(\na\) \\
                    Customer \#2 & 28 & 60000 & Teacher & Los Angeles & \(\prec\) & & \(\na\) \\
                    Customer \#3 & 42 & 65000 & Lawyer & Chicago & \(\sim\) & & \(\sim\) \\
                    Customer \#4 & 30 & 45000 & Doctor & Miami & \(\succ\) & & \(\succ\) \\
                    \hline
                \end{tabu}
            }
        };
    
        \node (classifier) [function, below = 3cm of datasets] {\(\cM\)};
        \path (datasets) edge node [annotation, above, sloped] {Trained} (classifier);
        \node (information) [annotation, below = 0.1cm of classifier] {Classifier};
    
        \node (instance) [instance, right = 2cm of classifier] {Instance};
        \path (instance) edge node [annotation, above] {Given to} (classifier);
        \node [above = 0.25cm of instance]
        {
            \scalebox{0.5}
            {
                \centering
                \begin{tabu}{|rrrr|}
                    \hline
                    Income & Age & Occupation & City \\
                    \hline
                    35000 & 37 & Teacher & Chicago \\
                    \hline 
                \end{tabu}
            }
        };
    
        \node (probabilities) [preference, left = 2cm of classifier] {Posterior\\probabilities};
        \path (classifier) edge node [annotation, above] {Predict} (probabilities);
        \node [above = 0.25cm of probabilities]
        {
            \scalebox{0.5}
            {
                \centering
                \begin{tabu}{|l|rrr|}
                    \hline
                    Pairwise variables & \(\succ\) & \(\sim\) & \(\prec\) \\
                    \hline
                    Hawaii and Bali & 0.5 & 0.25 & 0.25 \\
                    \hline
                    \multicolumn{4}{|c|}{\dots} \\
                    \hline
                    Tokyo and Miami & 0.125 & 0.25 & 0 \\
                    \hline
                \end{tabu}
            }
        };
    
        \node (matrix) [preference, left = 2cm of probabilities] {Preference\\matrix};
        \path (probabilities) edge node [annotation, above] {Filled} (matrix);
        \node [above = 0.25cm of matrix]
        {
            \scalebox{0.5}
            {
                \centering
                \begin{tabu}{l|rrrrr|}
                    \tabucline{2-6}
                    & Hawaii & Bali & Paris & Miami & Tokyo \\
                    \hline
                    \multicolumn{1}{|l|}{Hawaii} & 0.5 & 0.625 & 0.75 & 0.875 & 0.75 \\
                    \multicolumn{1}{|l|}{Bali} & 0.375 & 0.5 & 0.5 & 0.875 & 0.5 \\
                    \multicolumn{1}{|l|}{Paris} & 0.25 & 0.5 & 0.5 & 0.75 & 0.5 \\
                    \multicolumn{1}{|l|}{Miami} & 0.125 & 0.125 & 0.25 & 0.5 & 0.25 \\
                    \multicolumn{1}{|l|}{Tokyo} & 0.25 & 0.5 & 0.5 & 0.75 & 0.5 \\
                    \hline
                \end{tabu}
            }
        };

        \node (obop) [function, left = 2cm of matrix] {\(\mathcal{A}\)};
        \path (matrix) edge node [annotation, above, sloped] {Provided to} (obop);
        \node [annotation, below = 0.1cm of obop] {Optimal bucket\\order problem\\algorithm};
    
        \node (prediction) [prediction, left = 2cm of obop] {\(\widehat{\B}\)};
        \path (obop) edge node [annotation, above] {Aggregates} (prediction);
        \node [annotation, below = 0.1cm of prediction] {Prediction};
        \node [above = 0.25cm of prediction]
        {
            \scalebox{0.5}
            {%
                \centering
                \begin{tabu}{|r|}
                    \hline
                    Vacation destination \\
                    \hline
                    Bali \(\sim\) Miami \(\succ\) Tokyo \(\sim\) Paris \(\sim\) Hawaii \\
                    \hline
                \end{tabu}
            }
        };

    \end{tikzpicture}
}
        \caption{PLR}
        \label{figure:plr}
    \end{subfigure}%
    \vspace{0.5cm}
    \begin{subfigure}{\textwidth}
        \resizebox{\linewidth}{!}
{
    \begin{tikzpicture}[thick]

        \setlength{\tabcolsep}{0.15cm}
    
        \node (dataset) [dataset] {Dataset};
        \node [above = 0.25 of dataset]
        {
            \scalebox{0.5}
            {
                \centering
                \begin{tabu}{|lrrrr|r|}
                    \hline
                    Instance & Age & Income & Occupation & City & Vacation destination \\
                    \hline
                    Customer \#1 & 35 & 50000 & Engineer & New York & Hawaii \(\succ\) Bali \(\sim\) Paris \(\succ\) Miami \\
                    Customer \#2 & 28 & 60000 & Teacher & Los Angeles & Bali \(\succ\) Hawaii \(\succ\) Miami \\
                    Customer \#3 & 42 & 65000 & Lawyer & Chicago & Paris \(\sim\) Bali \(\sim\) Hawaii \(\sim\) Tokyo \(\sim\) Miami \\
                    Customer \#4 & 30 & 45000 & Doctor & Miami & Hawaii \(\succ\) Bali \(\sim\) Tokyo \(\succ\) Miami \\
                    \hline
                \end{tabu}
            }
        };

        \node (encoder) [function, right = 3.25cm of dataset] {\(\lambda\)};
        \path (dataset) edge node [annotation, above] {Used by} (encoder);
        \node [annotation, below = 0.1cm of encoder] {Encoder};
    
        \node (datasets) [dataset, right = 3.25cm of encoder] {Datasets};
        \path (encoder) edge node [annotation, above] {Transforms to} (datasets);
        \node [above = 0.25 of datasets]
        {
            \scalebox{0.5}
            {
                \centering
                \centering
                \begin{tabu}{|lrrrr|r|r|r|r|r|}
                    \hline
                    Instance & Age & Income & Occupation & City & Hawaii & Bali & Paris & Tokyo & Miami \\
                    \hline
                    Customer \#1 & 35 & 50000 & Engineer & New York & 1 & 2 & 2 & \(\na\) & 4 \\
                    Customer \#2 & 28 & 60000 & Teacher & Los Angeles & 2 & 1 & \(\na\) & \(\na\) & 3 \\
                    Customer \#3 & 42 & 65000 & Lawyer & Chicago & 1 & 1 & 1 & 1 & 1 \\
                    Customer \#4 & 30 & 45000 & Doctor & Miami & 1 & 2 & \(\na\) & 2 & 4 \\
                    \hline
                \end{tabu}
            }
        };

        \node (regressor) [function, below = 3cm of datasets] {\(\mathcal{H}\)};
        \path (datasets) edge node [annotation, above, sloped] {Trained} (regressor);
        \node [annotation, below = 0.1cm of regressor] {Multi-output regressor};
    
        \node (instance) [instance, right = 2cm of regressor] {Instance};
        \path (instance) edge node [annotation, above] {Given to} (regressor);
        \node [above = 0.25cm of instance]
        {
            \scalebox{0.5}
            {
                \centering
                \begin{tabu}{|rrrr|}
                    \hline
                    Income & Age & Occupation & City \\
                    \hline
                    35000 & 37 & Teacher & Chicago \\
                    \hline 
                \end{tabu}
            }
        };
    
        \node (values) [preference, left = 2cm of regressor] {Posterior\\values};
        \path (regressor) edge node [annotation, above] {Predicts} (values);
        \node [above = 0.25cm of values]
        {
            \scalebox{0.5}
            {
                \centering
                \begin{tabu}{|r|r|r|r|r|}
                    \hline
                    Hawaii & Bali & Paris & Tokyo & Miami \\
                    \hline
                    1.2 & 2.2 & 1.5 & 1.3 & 2.8 \\
                    \hline
                \end{tabu}
            }
        };

        \node (layer) [function, left = 2cm of values] {\(\sigma\)};
        \path (values) edge node [annotation, above, sloped] {Provided to} (layer);
        \node [annotation, below = 0.1cm of layer] {Post-hoc layer};
    
        \node (prediction) [prediction, left = 2cm of layer] {\(\widehat{\B}\)};
        \path (layer) edge node [annotation, above] {Transforms} (prediction);
        \node [annotation, below = 0.1cm of prediction] {Prediction};
        \node [above = 0.25cm of prediction]
        {
            \scalebox{0.5}
            {%
                \centering
                \begin{tabu}{|r|}
                    \hline
                    Vacation destination \\
                    \hline
                    Hawaii \(\sim\) Tokyo \(\succ\) Paris \(\succ\) Bali \(\succ\) Miami \\
                    \hline
                \end{tabu}
            }
        };
    \end{tikzpicture}
}%
        \caption{MORE}
        \label{figure:more}
    \end{subfigure}
    \caption{In this example, the goal is to predict a bucket order of vacation destinations for a customer based on their demographic information.}
\end{figure}

    This section presents our meta-learning approach for the \gls{plr} problem.
The first step is to transform the bucket orders in the training dataset into a suitable representation for a \gls{mor} model.
This representation is by means of a \textit{rank position vector} derived by an \textit{encoder}, which results in a well-defined training dataset for the \gls{mor} model.
At the inference phase, the \gls{mor} model predicts a rank position vector for a given input instance.
However, the predicted rank position vector may not directly correspond to a bucket order, so we employ a \gls{plr} post-hoc layer to transform it accordingly.
At a high level, the \gls{more-plr} framework (see Fig. \ref{figure:more}) mirrors the standard \gls{plr} learning and inference processes (see Fig. \ref{figure:plr}). Specifically, the \textit{multi-output regressor} takes the role of the \textit{classifier}. At the same time, the \gls{plr} post-hoc layer replaces the \gls{obop} step.
In both frameworks, there is also a step that converts (or encodes) the original dataset into a new representation.
However, the transformation or encoding component in the learning process of \gls{plr} is fixed (or does not exist), as a direct conversion to classification datasets is always considered.
Our extended framework\footnote{Extended compared to the previous version of the paper as explained at the end of Section \ref{section:introduction}.}, however, now provides multiple ways to generate a meaningful dataset for the \gls{mor} component. 

        \subsection{Bucket Orders as Rank Position Vectors}
        \label{subsection:encoding}
        Currently, the basic building block for learning in the \gls{plr} problem is that a bucket order \(\B\) is equivalently represented by bucket matrix through \eqref{def:bucket_matrix}.
However, another equivalent representation of a bucket order \(\B\) is a \textit{rank position vector} \(\vec{p} = (p_1, \ldots, p_k)\), which captures the ranking relationships among the items in \(\mathcal{I}\).
Different encoding schemes can construct a rank position vector, each providing a distinct but equivalent perspective on the bucket order structure. These methods are:
\begin{itemize}
    \item \textit{dense}, given by
        \begin{align}
        \label{equation:dense}
                p_i = 1 + \sum\nolimits_{l = 1}^c \id_{\{\bucket{l} \succ_{\buckets} u_i\}}, \quad i \in [k],
        \end{align}
        where \(\id_{\{\cdot\}}\) is the indicator function and \(\bucket{l} \succ_{\buckets} u_i\) means \(u_{i}\) is in a subsequent bucket of \(\bucket{l}\);
    \item \textit{standard competition}, given by
        \begin{align}
        \label{equation:standard}
            p_i = 1 + \sum\nolimits_{l = 1}^c \id_{\{\bucket{l} \succ_{\buckets} u_i\}} |\bucket{l}| , \quad i \in [k],
        \end{align}
        where \(|\bucket{l}|\) denotes the cardinality of a set \(\bucket{l} \subseteq \mathcal{I}\);
    \item \textit{modified competition}, given by
        \begin{align}
        \label{equation:modified}
            p_i = \sum\nolimits_{l = 1}^c \id_{\{\bucket{l} \succ_{\buckets} u_i \}} |\bucket{l}| + \id_{\{u_i \in \bucket{l}\}} |\bucket{l}|, \quad i \in [k];
        \end{align}
    \item and \textit{fractional}, given by
        \begin{align}
        \label{equation:fractional}
            p_i = \sum\nolimits_{l = 1}^c \id_{\{\bucket{l} \succ_{\buckets} u_i\}} |\bucket{l}| + \id_{\{u_i \in \bucket{l}\}} \big(1 + (|\bucket{l}| - 1) \cdot 0.5 \big), \quad i \in [k].
        \end{align}
\end{itemize}

For illustration, consider a case where we have \(k = 4\) and the bucket order \(\B = \langle\{u_1\}, \{u_2, u_3\}, \{u_4\}\rangle\).
This means that \(u_1\) is ranked highest, followed by \(u_2\) and \(u_3\) (which share the same rank), and finally \(u_4\), which has the lowest rank.
We then obtain (see Appendix~\ref{appendix:encoding} for a detailed calculation):%
\begin{itemize}
    \item \(\vec{p} = (1, 2, 2, 3)\) using \eqref{equation:dense},
    \item \(\vec{p} = (1, 2, 2, 4)\) using \eqref{equation:standard},
    \item \(\vec{p} = (1, 3, 3, 4)\) using \eqref{equation:modified},
    \item and \(\vec{p} = (1, 2.5, 2.5, 4)\) using \eqref{equation:fractional}.
\end{itemize}

With a \gls{plr} dataset consisting only of complete rankings \(\cD \subseteq \cX \times \mathcal{S}_k^{\succeq}\), learning a multi-output regressor is straightforward, as the \gls{mor} dataset \(\cD \subseteq \cX \times \mathbb{R}^d\) provides a fully defined multivariate target for each instance.
However, when transforming a \gls{plr} dataset with (possibly incomplete) rankings with ties into a \gls{mor} dataset \(\cD \subseteq \cX \times (\R \cup \{\na\})^d\), missing labels are assigned a \(\na\) value into the rank positions (see Fig. \ref{figure:more}).
Therefore, the multi-output regressor must handle these missing values appropriately.
The simplest approach is to discard all data points containing at least one \(\na\) value from the dataset and train the \gls{mor} model on the fully observed data.
However, this can significantly reduce the available training data, potentially harming model performance.
A more effective strategy depends on the choice of \gls{mor} method.
For instance, in the single-target approach, each target variable is trained independently on all data points where its value is available.
This allows the model to use a more extensive training dataset than the naive approach of removing all incomplete samples.
As shown in Section \ref{section:experiments}, this single-target approach improves performance when handling incomplete data by maximizing the use of available information while minimizing the negative impact of missing values.

Finally, note that all encoding variants yield the same results if the bucket order is a total order.
This is also reflected in our experiments, which we will comment on at the relevant point.

        \subsection{PLR post-hoc Layers}
        \label{subsection:layers}
        Predictions in \gls{plr} problems are bucket orders of the underlying items, which can be encoded equivalently using a bucket order matrix or a rank position vector.
However, the output from the multi-output regressor does not directly yield a rank position vector that accurately represents a bucket order.
Instead, the predictions are usually \(k\)-dimensional vectors, which we refer to as prediction vectors moving forward.
Therefore, these prediction vectors must be transformed into a bucket order.

To achieve this, we introduce several transformations, described below, which can be viewed as post-hoc layers within the overall inference phase.
These layers, termed \textit{\gls{plr} post-hoc layers}, are mappings of the form \(\sigma: \R^k \rightarrow \mathcal{S}_k^{\succeq}\) and operate on the output \(\vec{\haty} = H(\x)\) of a \gls{mor} model (see Fig. \ref{figure:more}).
The core of these layers is the construction of bucket order \(\succeq_{\mathcal{B}(\sigma)}\) for the entries of the output vector \(\vec{\haty}\). 

\subsubsection*{\Gls{rr} Layer}

The bucket order \(\succeq_{\mathcal{B}(\sigma_{\mathrm{\gls{rr}}})}\) is determined from entries of output vector \(\vec{\haty}\) as follows:
\begin{align*}
    &\haty_i \succ_{\mathcal{B}(\sigma_{\mathrm{\gls{rr}}})}  \haty_j \quad \Leftrightarrow \quad \round(\haty_i) > \round(\haty_j), \\
    &\haty_i \sim_{\mathcal{B}(\sigma_{\mathrm{\gls{rr}}})}  \haty_j \quad \Leftrightarrow \quad \round(\haty_i) = \round(\haty_j),
\end{align*}
where \(\round(x)=\argmin_{n\in \N}|n-x|\) is the rounding operator\footnote{We take the larger number in the case of two minimizers.}.

\subsubsection*{\Gls{pi} Layer}

This layer is based on the concept of overlapping intervals.
If we have prediction intervals of the form \([\haty_i - c_i, \haty_i + c_i]\) available, we can construct a bucket order \(\succeq_{\mathcal{B}(\sigma_{\mathrm{\gls{pi}}})}\) as follows:
\begin{align*}
    \haty_i \succ_{\mathcal{B}(\sigma_{\mathrm{\gls{pi}}})} \haty_j  
    \quad \Leftrightarrow \quad &\nexists \{i_1, \ldots, i_l\} \subset \{1, \ldots, k\} \setminus\{i, j\}: \  \forall r = 0, \ldots, l, \\ 
    &[\hat{y}_{i_r} - c_{i_r}, \hat{y}_{i_r} + c_{i_r}] \cap [\hat{y}_{i_{r + 1}} - c_{i_{r + 1}}, \hat{y}_{i_{r + 1}} + c_{i_{r + 1}}] \neq \emptyset, \\
    \haty_i \sim_{\mathcal{B}(\sigma_{\mathrm{\gls{pi}}})} \haty_j 
    \quad \Leftrightarrow \quad &\exists \{i_1, \ldots, i_l\} \subset \{1, \ldots, k\} \setminus\{i, j\}: \ \forall r = 0, \ldots, l, \\ 
    &[\hat{y}_{i_r} - c_{i_r}, \hat{y}_{i_r} + c_{i_r}] \cap [\hat{y}_{i_{r + 1}} - c_{i_{r + 1}}, \hat{y}_{i_{r + 1}} + c_{i_{r + 1}}] \neq \emptyset,  
\end{align*}
where we set \(i_0 = i\) and \(i_{l + 1} = j\). 
In other words, items are placed in the same bucket if their prediction intervals overlap or if there is a cascading effect: if a position \(i\) overlaps with position \(j\) and this overlaps extends to a third position \(m\), then \(i\) and \(m\) get assigned the same bucket due to their overlap with \(j\).

It is worth noting that this can lead to an undesired bucket order \(\succeq_{\mathcal{B}(\sigma_{\mathrm{\gls{pi}}})}\) representing essentially only one bucket. 
However, it offers the advantage of being grounded in the models' uncertainty when multiple items are placed in the same bucket.
In fact, the prediction interval represents all possible predictions given the model's uncertainty. 
If these intervals overlap, it suggests that the items have the same rank from the model's perspective of uncertainty.
Unfortunately, not all methods inherently provide prediction intervals, but standard techniques such as \textit{Gaussian processes} or \textit{random forests} do.
Alternatively, using conformal prediction, one can construct prediction intervals for any prediction method \parencite{angelopoulos_conformal_2023}.

\subsubsection*{\Gls{eps} Layer}

Another transformation can be obtained by first normalizing the output to be in \([0, 1]^k\) and then giving all items that are \(\epsilon\)-close to each other the same rank, where \(\epsilon \in (0, 1)\) is some small value.
Let
\begin{equation*}
\norm(\vec{x}) = \Big(\tfrac{x_1 - \min_j x_j}{\max_j x_j - \min_j x_j},\ldots, \tfrac{x_k - \min_j x_j}{\max_j x_j - \min_j x_j}\Big)
\end{equation*}
be the min-max normalization of a vector \(\vec{x} \in \R^k\), then, the bucket order \(\succeq_{\mathcal{B}(\sigma_{\mathrm{EPS}})}\) is obtained form the entries of the output vector \(\vec{\haty}\) as follows:
\begin{align*}
    \haty_i \succ_{\mathcal{B}(\sigma_{\mathrm{EPS}})} \haty_j  
    \quad \Leftrightarrow \quad &\nexists \{i_1, \ldots, i_l\} \subset \{1, \ldots, k\} \setminus\{i, j\} \ \mbox{such that} \\ 
    &|\norm(\vec{\haty})_{i_r} - \norm(\vec{\haty})_{i_{r + 1}}|< \epsilon, r = 0, \ldots, l,  \\
    \haty_i \sim_{\mathcal{B}(\sigma_{\mathrm{EPS}})} \haty_j 
    \quad \Leftrightarrow \quad &\exists \{i_1, \ldots, i_l\} \subset \{1, \ldots, k\} \setminus\{i, j\} \ \mbox{such that} \\ 
    &|\norm(\vec{\haty})_{i_r} - \norm(\vec{\haty})_{i_{r + 1}}|< \epsilon, r = 0,\ldots, l, 
\end{align*}
where we set \(i_0 = i\) and \(i_{l + 1} = j\).

Note that, as described above for the \gls{pi} layer, the construction of \(\succeq_{\mathcal{B}(\sigma_{\mathrm{EPS}})}\) also accounts for a cascading effect when determining equal positions.

        \section{Experiments}
        \label{section:experiments}
        This section investigates the performance of some learning approaches in the \gls{more-plr} class and compares them to the current state-of-the-art models for the \gls{plr} problem.
We first describe the technical background of the experimental evaluation and then present the results on the standard benchmarking datasets.

            \subsection{Methodology}
            \label{subsection:methodology}
            The base models did not receive significant fine-tuning of the hyperparameters, and the results can be seen as out-of-the-box performance.
To compensate for potential random deviations, all algorithms were evaluated using five repetitions of a ten-fold cross-validation method.
We measured the accuracy using the \(\tau_X\) \textit{rank correlation coefficient} \parencite{emond_new_2002}. 
Formally, given two bucket orders \(\succeq_{\mathcal{B}^r}\) and \(\succeq_{\mathcal{B}^p}\) defined over the items \( \mathcal{I}\), the \(\tau_X\) rank correlation coefficient is given by
\begin{equation}
	 \tau_X(\succeq_{\mathcal{B}^r}, \succeq_{\mathcal{B}^p}) = \tfrac{\sum_{i = 1}^{k} \sum_{j = 1}^k \beta_{ij}^r \beta_{ij}^p}{k (k - 1)},
\end{equation}
\noindent where
\[
    \beta_{ij}^l = \begin{cases}
                            1 & \text{ if } u_i \succ_{\mathcal{B}^l} u_j \text{ or } u_i \sim_{\mathcal{B}^l} u_j, \\
                            -1 & \text{ if } u_j \succ_{\mathcal{B}^l} u_i, \\
                            0 & \text{ if } i = j.
                    \end{cases}
\]
The \(\tau_X\) rank correlation coefficient values lie in the interval \([-1, 1]\). Hereby, \(1\) indicates a perfect correlation between the bucket orders, \(-1\) is a perfect correlation of a bucket order to its reversed bucket order, and a value close to zero with no correlation. Thus, the model aims to achieve values of \(\tau_X\) to \(1\).
When speaking of CPU time in the following, we mean the time needed to train a learner and the time needed for inference averaged over the \(5 \times 10\)-cv.

The results were analyzed using the procedure of \textcite{demsar_statistical_2006, garcia_extension_2008} with the \texttt{exreport} software tool \parencite{arias_exreport:_2015}.
First, a \textit{Friedman test} \parencite{friedman_comparison_1940} was applied with the null hypothesis that all the algorithms have equal performance.
If this hypothesis was rejected, a \textit{post-hoc test} using \textit{Holm's procedure} \parencite{holm_simple_1979} was performed to compare all the algorithms against the one ranked first by the Friedman test. Both tests were conducted at a significance level of 5\%.

We also considered  \(30\%\) and \(60\%\) of missing labels, which are removed from the training datasets according to the standard procedure described in \textcite{cheng_decision_2009, alfaro_learning_2021}.

            \subsection{Datasets and Algorithms}
            \label{section:algorithms}
            We used the standard datasets on which methods for the \gls{plr} problem have been experimentally investigated \parencite{alfaro_learning_2021, alfaro_ensemble_2022, alfaro_pairwise_2023}.
These are modifications of datasets used for \gls{lr} \parencite{cheng_decision_2009}, which we also included to investigate potential differences in performance for these two settings.
Appendix \ref{appendix:datasets} provides a detailed description of the datasets.

The below-listed algorithms were used:

\begin{itemize}
    \item \textit{Pairwise comparison reduction} technique \parencite{alfaro_pairwise_2023}. It is the current state-of-the-art method for the \gls{plr} problem, called \gls{rpc}.
    \item \textit{Single-target} method. This learner applies the \gls{st} method as \gls{mor}~technique. 
    \item \textit{Chain} method. We also investigate the chaining (Chain) method as the underlying \gls{mor} method, which utilizes a possible dependency between the targets, which we call \textit{order}.
    However, to determine the best order, one must check all possible \(k!\) orders, which is not feasible.
    Therefore, we propose using the correlation of the target variables as a good indicator. To build an order, we first look at the target variable with the highest additive correlation compared to all other target variables and choose it as the starting target. This approach is applied recursively to the remaining targets until the order is created.
    \item \textit{Algorithm adaption} method. Since \textit{random forest} regressors can directly model \gls{mor} \parencite{borchani_survey_2015}, we also included it as a representative of the algorithm adaption (Native) methods.
\end{itemize}
We used \textit{random forests} (classifier or regressor, as corresponding) as the base method for all the algorithms.
Random forests, as an ensemble method, allows for obtaining prediction intervals as follows: if \(\#trees\) is the number of trees and \(\sigma_i\) is the standard deviation of the individual outputs, we set
\begin{equation*}
    c_i = \frac{q\sigma_i}{\sqrt{\#trees}},
\end{equation*}
which corresponds to the sample standard error.
The \(\sigma\)-factor \(1 \leq q \leq 3\) specifies the probability that the confidence interval will overlap with the true target corresponding to the \(3\sigma\) rule.
Note that this could be adapted by multiplying it by a quantile of the normal distribution, but we do not do it for the sake of simplicity.

\begin{figure}
    \centering
    \begin{subfigure}{\textwidth}
        \includegraphics[width = \textwidth]{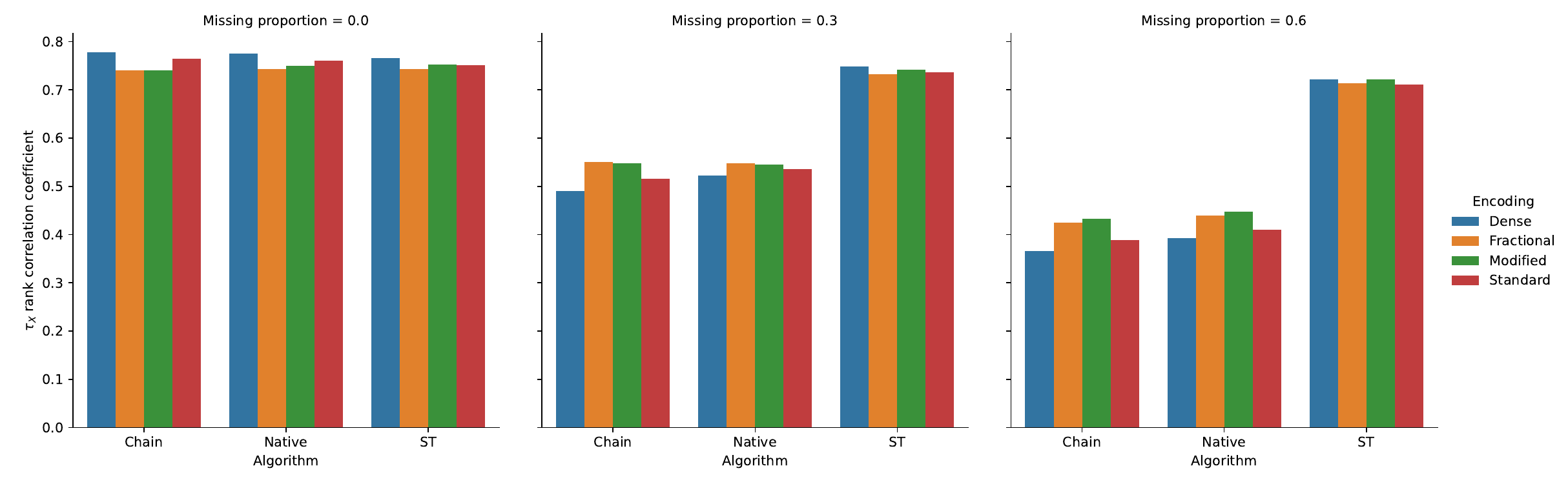}
        \caption{\gls{rr} layer}
        \label{figure:rr_layer}
    \end{subfigure}%
    \vspace{0.25cm}
    \begin{subfigure}{\textwidth}
        \includegraphics[width = \textwidth]{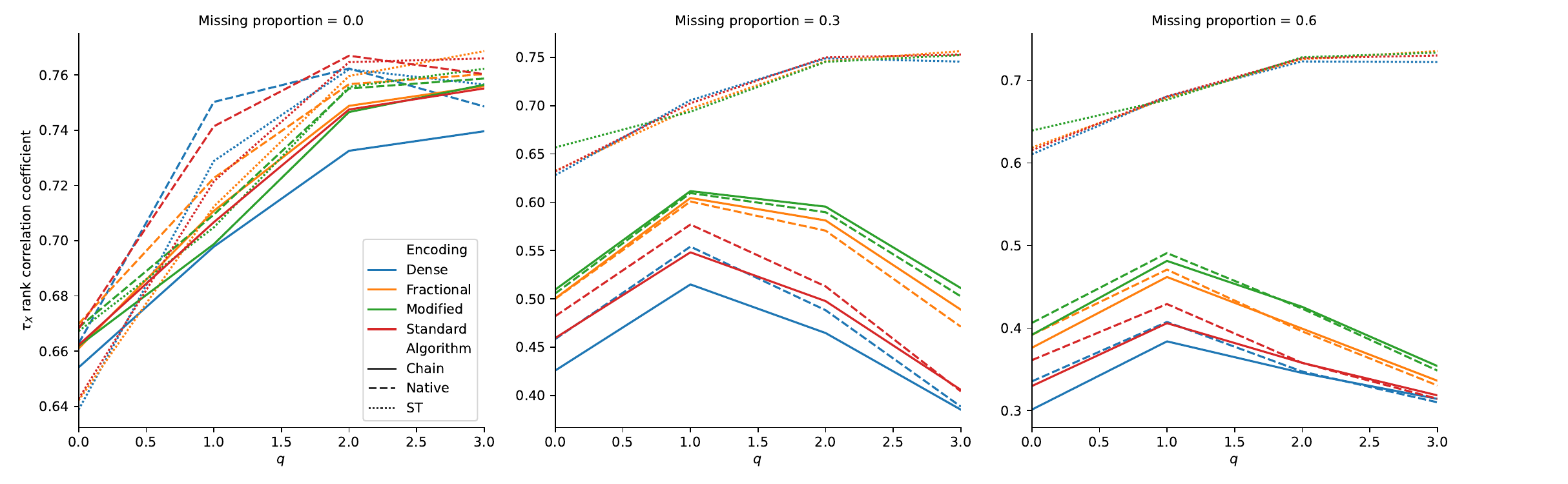}
        \caption{\gls{pi} layer}
        \label{figure:pi_layer}
    \end{subfigure}%
    \vspace{0.25cm}
    \begin{subfigure}{\textwidth}
        \includegraphics[width = \textwidth]{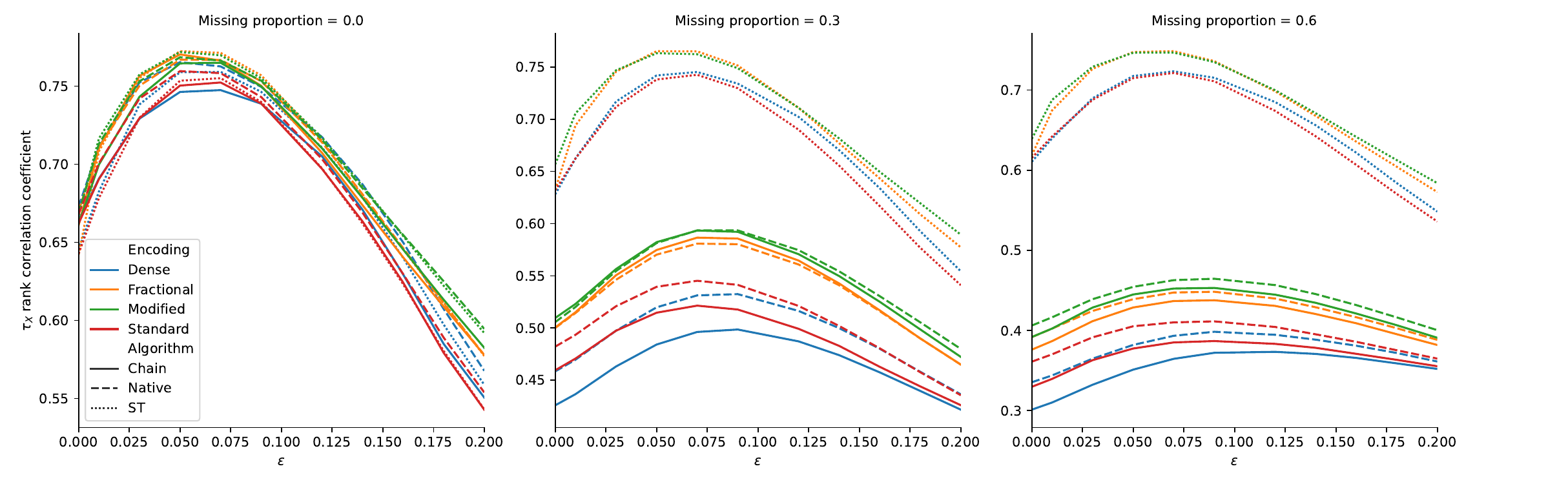}
        \caption{\gls{eps} layer}
        \label{figure:eps_layer}
    \end{subfigure}
    \caption{Impact of encoding variants on algorithms' averaged accuracy across all datasets}
    \label{figure:encoding}
\end{figure}

All algorithms were tested with every possible \gls{plr} post-hoc layer and encoding variant for a thorough evaluation.
However, given the large number of hyperparameter combinations available, our first goal is to narrow them down and retain only the most promising algorithms for the statistical tests.
To achieve this, we first determine which encoding variant is more suitable for each \gls{plr} post-hoc layer.
Fig.~\ref{figure:encoding} shows the averaged accuracy results\footnote{Only \gls{plr} problems were tested, as the encoding variants have no effect on \gls{lr} problems.} over all \gls{plr} datasets, leading to the following conclusions:

\begin{itemize}

    \item \gls{rr} layer. At 0\% missing positions, the dense encoding performed best across all methods (Chain, Native, and \gls{st}). When 30\% of positions were missing, the fractional ranking was the most suitable for Chain and Native, while ST continued to perform best with the dense encoding. At 60\% missing positions, the modified competition encoding was preferred for Chain and Native, whereas ST still achieved the highest performance with the dense encoding. Since \gls{st} is the most promising algorithm and performs best with the dense encoding, we selected it as the general choice for the \gls{rr} layer.

    \item \gls{pi} layer. At 30\% and 60\% of positions were missing, the modified competition encoding was the most suitable across all methods and for varying values of \(q\). At 0\% missing positions, there was no clear best-performing encoding. However, since the modified competition encoding consistently performed best at higher missing proportions, we selected it for consistency, as the general choice for the \gls{pi} layer.

    \item \gls{eps} layer. Regardless of the percentage of missing positions or the \(\epsilon\) value, the modified competition encoding consistently outperformed other approaches across all methods. Given its stability and effectiveness in all scenarios, we selected the modified competition encoding as the general choice for the \gls{eps} layer.

\end{itemize}

\begin{figure}[t]
    \centering
    \includegraphics[width = \textwidth]{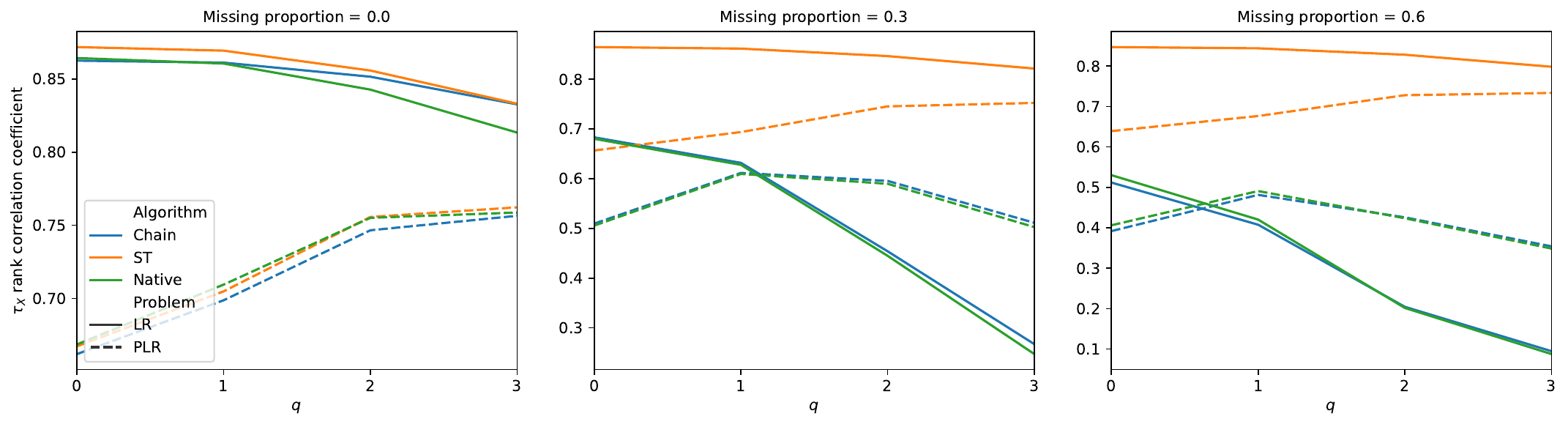}
    \caption{Impact of \(q\) in the accuracy of the algorithms}
    \label{figure:q_values}
\end{figure}

The next step is to select an appropriate value for \(q\) in the \gls{pi} layer, as it impacts the accuracy of the algorithms. Fig. \ref{figure:q_values} illustrates the effect of different values.
According to these results, for the \gls{lr} problem, lower values of \(q\) improve accuracy because the prediction contains a higher number of buckets \(c\), making it more similar to a total order.
On the other hand, for the \gls{plr} problem, higher values of \(q\) can positively impact the accuracy up to a certain point, as this leads to fewer buckets \(c\) in the prediction, increasing the likelihood of generating a bucket order with \(c \ll k\).
Therefore, to balance across both problems, we used \(q = 1\).

\begin{figure}
    \centering
    \includegraphics[width = \textwidth]{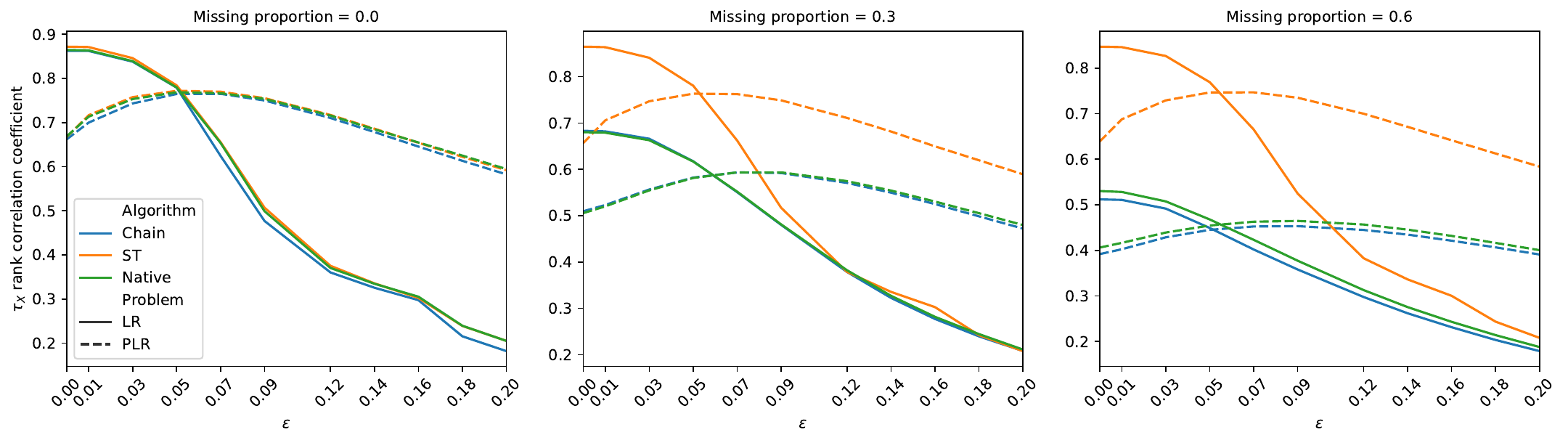}
    \caption{Impact of \(\epsilon\) on the accuracy of the algorithms}
    \label{figure:epsilon_values}
\end{figure}

The following step is to choose a proper \(\epsilon\) value in the \gls{eps} layer, as it affects in the accuracy of the algorithms.
As shown in Fig.~\ref{figure:epsilon_values}, the best value for the \gls{lr} problem is 0, while for the \gls{plr} problem, the optimal range lies between 0.05 and 0.07, regardless of the proportion of missing labels.
This can be explained by the fact that lower values of \(\epsilon\) tend to produce a total order, which typically improves the accuracy in \gls{lr} problems.
In constrast, higher values of \(\epsilon\) generate fewer buckets, which can enhance performance in \gls{plr} problems.
However, this comes with the risk of producing too few buckets, reducing the model's ability to accurately match predictions to the ground truth and potentially lowering accuracy.
To select a value that works well for both problems, we chose \(\epsilon = 0.03\).

Based on this preliminary analysis, we selected for the statistical test:
\begin{itemize}
    \item \gls{rr} layer with dense encoding;
    \item \gls{pi} layer with modified competition encoding and \(q = 1\);
    \item \gls{eps} layer with modified competition encoding and \(\epsilon = 0.03\).
\end{itemize}

Finally, it is important to note that selecting \(q\) and \(\epsilon\) values specifically tailored to each problem type could lead to improved results.

            \subsection{Results}
            \label{section:results}
            This section analyzes the algorithms regarding accuracy and efficiency.

\subsubsection*{Accuracy}

Table \ref{table:tests} summarizes the hypothesis tests conducted, organized by problem tested (columns) and percentage of missing labels (rows).
Appendix \ref{appendix:results} contains the complete results.
Both Friedman's test \(p\)-value and Holm's post-hoc test results are displayed for each group.  
It should be noted that even if the dataset consists only of total orders, that is, if the problem is a \gls{lr} problem, \gls{plr} estimators may still predict bucket orders.

\begin{table*}[t]
    \setlength{\tabcolsep}{0.2cm}
    \centering
    \caption{Friedman's and Holm's tests for accuracy with varying missing percentage}
    \resizebox{\textwidth}{!}
    {
        \begin{tabu}{|[0.1cm]lrrrrr|}
            \tabucline[0.1cm]{-}
            \multicolumn{6}{|[0.1cm]c|}{\textbf{\gls{lr}}} \\
            \tabucline[0.1cm]{-}
            \multicolumn{6}{|[0.1cm]c|}{\textbf{Missing percentage: 0\%}} \\
            \hline
            \multicolumn{6}{|[0.1cm]c|}{Friedman p-value: \(\bf 5.136 \times 10^{-7}\)} \\
            \hline
            \multicolumn{6}{|[0.1cm]c|}{Holm results} \\
            \hline
            Method & Rank & P-value & Win & Tie & Loss \\
            \hline
            \gls{st}-\gls{pi} & 2.27 & - & - & - & -\\
            Native-\gls{pi} & 4.42 & \(\bf 1.146 \times 10^{-1}\) & 10 & 0 & 3 \\
            \gls{st}-\gls{eps} & 4.54 & \bf \(\bf 1.146 \times 10^{-1}\) & 9 & 1 & 3 \\
            Chain-\gls{pi} & 4.73 & \bf \(\bf 1.146 \times 10^{-1}\) & 12 & 0 & 1 \\
            \gls{st}-\gls{rr} & 4.96 & \bf \(\bf 9.353 \times 10^{-2}\) & 12 & 0 & 1 \\
            Native-\gls{eps} & 5.65 & \(2.185 \times 10^{-2}\) & 10 & 0 & 3 \\
            Native-\gls{rr} & 6.12 & \(7.203 \times 10^{-3}\) & 10 & 0 & 3 \\
            Chain-\gls{eps} & 6.23 & \(5.951 \times 10^{-3}\) & 12 & 0 & 1 \\
            Chain-\gls{rr} & 6.38 & \(4.235 \times 10^{-3}\) & 12 & 0 & 1 \\
            \gls{rpc} & 9.69 & \(3.675 \times 10^{-9}\) & 13 & 0 & 0 \\
            \tabucline[0.1cm]{-}
            \multicolumn{6}{|[0.1cm]c|}{\textbf{Miss percentage: 30\%}} \\
            \hline
            \multicolumn{6}{|[0.1cm]c|}{Friedman p-value: \(\bf 1.489 \times 10^{-15}\)} \\
            \hline
            \multicolumn{6}{|[0.1cm]c|}{Holm results} \\
            \hline
            Method & Rank & P-value & Win & Tie & Loss \\
            \hline
            \gls{st}-\gls{pi} & 1.35 & - & - & - & -\\  
            \gls{st}-\gls{eps} & 2.35 & \(\bf 6.952 \times 10^{-1}\) & 9 & 1 & 3 \\  
            \gls{st}-\gls{rr} & 2.46 & \(\bf 6.952 \times 10^{-1}\) & 12 & 0 & 1 \\  
            \gls{rpc} & 4.85 & \(9.618 \times 10^{-3}\) & 13 & 0 & 0 \\  
            Chain-\gls{eps} & 5.88 & \(5.301 \times 10^{-4}\) & 13 & 0 & 0 \\  
            Native-\gls{eps} & 5.96 & \(5.085 \times 10^{-4}\) & 13 & 0 & 0 \\  
            Chain-\gls{rr} & 7.50 & \(1.317 \times 10^{-6}\) & 13 & 0 & 0 \\  
            Native-\gls{rr} & 7.85 & \(3.089 \times 10^{-7}\) & 13 & 0 & 0 \\  
            Chain-\gls{pi} & 8.35 & \(3.006 \times 10^{-8}\) & 13 & 0 & 0 \\  
            Native-\gls{pi} & 8.46 & \(1.869 \times 10^{-8}\) & 13 & 0 & 0 \\  
            \tabucline[0.1cm]{-}
            \multicolumn{6}{|[0.1cm]c|}{\textbf{Miss percentage: 60\%}} \\
            \hline
            \multicolumn{6}{|[0.1cm]c|}{Friedman p-value: \(\bf 7.549 \times 10^{-18}\)} \\
            \hline
            \multicolumn{6}{|[0.1cm]c|}{Holm results} \\
            \hline
            Method & Rank & P-value & Win & Tie & Loss \\
            \hline
            \gls{st}-\gls{pi} & 1.35 & - & - & - & - \\
            \gls{st}-\gls{eps} & 2.35 & \(\bf 6.952 \times 10^{-1}\) & 9 & 1 & 3 \\
            \gls{st}-\gls{rr} & 2.46 & \(\bf 6.952 \times 10^{-1}\) & 12 & 0 & 1 \\
            \gls{rpc} & 4.85 & \(9.618 \times 10^{-3}\) & 13 & 0 & 0 \\
            Chain-\gls{eps} & 5.88 & \(5.301 \times 10^{-4}\) & 13 & 0 & 0 \\
            Native-\gls{eps} & 5.96 & \(5.085 \times 10^{-4}\) & 13 & 0 & 0 \\
            Chain-\gls{rr} & 7.50 & \(1.317 \times 10^{-6}\) & 13 & 0 & 0 \\
            Native-\gls{rr} & 7.85 & \(3.089 \times 10^{-7}\) & 13 & 0 & 0 \\
            Chain-\gls{pi} & 8.35 & \(3.006 \times 10^{-8}\) & 13 & 0 & 0 \\ 
            Native-\gls{pi} & 8.46 & \(1.869 \times 10^{-8}\) & 13 & 0 & 0 \\
            \tabucline[0.1cm]{-}
        \end{tabu}%
        \begin{tabu}{|[0.05cm]lrrrrr|}
            \tabucline[0.1cm]{-}
            \multicolumn{6}{|[0.05cm]c|}{\textbf{\gls{plr}}} \\
            \tabucline[0.1cm]{-}
            \multicolumn{6}{|[0.05cm]c|}{\textbf{Missing percentage: 0\%}} \\
            \hline
            \multicolumn{6}{|[0.05cm]c|}{Friedman p-value: \(\bf 1.475 \times 10^{-20}\)} \\
            \hline
            \multicolumn{6}{|[0.05cm]c|}{Holm results} \\
            \hline
            Method & Rank & P-value & Win & Tie & Loss \\
            \hline
            \gls{rpc} & 2.00 & - & - & - & - \\
            Native-\gls{rr} & 2.92 & \(\bf 5.912 \times 10^{-1}\) & 15 & 0 & 3 \\
            Chain-\gls{rr} & 3.06 & \(\bf 5.912 \times 10^{-1}\) & 14 & 1 & 3 \\
            \gls{st}-\gls{rr} & 4.25 & \(\bf 7.735 \times 10^{-2}\) & 16 & 0 & 2 \\
            \gls{st}-\gls{eps} & 4.89 & \(1.681 \times 10^{-2}\) & 16 & 0 & 2 \\
            Native-\gls{eps} & 5.44 & \(3.213 \times 10^{-3}\) & 16 & 0 & 2 \\
            Chain-\gls{eps} & 6.42 & \(7.242 \times 10^{-5}\) & 16 & 0 & 2 \\
            \gls{st}-\gls{pi} & 8.25 & \(4.134 \times 10^{-9}\) & 16 & 1 & 1 \\
            Native-\gls{pi} & 8.50 & \(9.518 \times 10^{-10}\) & 17 & 0 & 1 \\
            Chain-\gls{pi} & 9.28 & \(4.987 \times 10^{-12}\) & 17 & 0 & 1 \\
            \tabucline[0.1cm]{-}
            \multicolumn{6}{|[0.05cm]c|}{\textbf{Miss percentage: 30\%}} \\
            \hline
            \multicolumn{6}{|[0.05cm]c|}{Friedman p-value: \(\bf 2.163 \times 10^{-25}\)} \\
            \hline
            \multicolumn{6}{|[0.05cm]c|}{Holm results} \\
            \hline
            Method & Rank & P-value & Win & Tie & Loss \\
            \hline
            \gls{rpc} & 1.83 & - & - & - & - \\
            \gls{st}-\gls{rr} & 2.36 & \(\bf 1.000\) & 12 & 0 & 6 \\
            \gls{st}-\gls{eps} & 2.42 & \(\bf 1.000\) & 12 & 0 & 6 \\
            \gls{st}-\gls{pi} & 4.00 & \(\bf 9.541 \times 10^{-2}\) & 15 & 0 & 3 \\
            Native-\gls{pi} & 5.33 & \(2.097 \times 10^{-3}\) & 18 & 0 & 0 \\
            Chain-\gls{pi} & 5.58 & \(1.013 \times 10^{-3}\) & 18 & 0 & 0 \\
            Native-\gls{eps} & 7.64 & \(5.275 \times 10^{-8}\) & 18 & 0 & 0 \\
            Chain-\gls{eps} & 7.72 & \(3.763 \times 10^{-8}\) & 18 & 0 & 0 \\
            Native-\gls{rr} & 8.50 & \(3.163 \times 10^{-10}\) & 18 & 0 & 0 \\
            Chain-\gls{rr} & 9.61 & \(1.162 \times 10^{-13}\) & 18 & 0 & 0 \\
            \tabucline[0.1cm]{-}
            \multicolumn{6}{|[0.05cm]c|}{\textbf{Missing percentage: 60\%}} \\
            \hline
            \multicolumn{6}{|[0.05cm]c|}{Friedman p-value: \(\bf  9.755 \times 10^{-25}\)} \\
            \hline
            \multicolumn{6}{|[0.05cm]c|}{Holm results} \\
            \hline
            Method & Rank & P-value & Win & Tie & Loss \\
            \hline
            \gls{st}-\gls{eps} & 2.00 & - & - & - & - \\
            \gls{st}-\gls{rr} & 2.17 & \(\bf 9.484 \times 10^{-1}\) & 11 & 0 & 7 \\
            \gls{rpc} & 2.72 & \(\bf 9.484 \times 10^{-1}\) & 12 & 0 & 6 \\
            \gls{st}-\gls{pi} & 3.61 & \(\bf 3.312 \times 10^{-1}\) & 17 & 0 & 1 \\
            Native-\gls{pi} & 5.50 & \(2.097 \times 10^{-3}\) & 17 & 0 & 1 \\
            Chain-\gls{pi} & 6.28 & \(1.124 \times 10^{-4}\) & 17 & 0 & 1 \\
            Native-\gls{eps} & 6.69 & \(1.977 \times 10^{-5}\) & 18 & 0 & 0 \\
            Chain-\gls{eps} & 7.72 & \(9.998 \times 10^{-8}\) & 18 & 0 & 0 \\
            Native-\gls{rr} & 8.61 & \(4.580 \times 10^{-10}\) & 17 & 0 & 1 \\
            Chain-\gls{rr} & 9.69 & \(2.210 \times 10^{-13}\) & 17 & 0 & 1 \\
            \tabucline[0.1cm]{-}
        \end{tabu}%
        \begin{tabu}{|[0.05cm]lrrrrr|[0.1cm]}
            \tabucline[0.1cm]{-}
            \multicolumn{6}{|[0.05cm]c|[0.1cm]}{\textbf{Both}} \\
            \tabucline[0.1cm]{-}
            \multicolumn{6}{|[0.05cm]c|[0.1cm]}{\textbf{Miss percentage: 0\%}} \\
            \hline
            \multicolumn{6}{|[0.05cm]c|[0.1cm]}{Friedman p-value: \(\bf 8.100 \times 10^{-5}\) } \\
            \hline
            \multicolumn{6}{|[0.05cm]c|[0.1cm]}{Holm results} \\
            \hline
            Method & Rank & P-value & Win & Tie & Loss \\
            \hline
            Native-\gls{rr} & 4.26 & - & - & - & - \\
            Chain-\gls{rr} & 4.45 & \(\bf 1.000\) & 12 & 1 & 18 \\
            \gls{st}-\gls{rr} & 4.55 & \(\bf 1.000\) & 18 & 3 & 10 \\
            \gls{st}-\gls{eps} & 4.74 & \(\bf 1.000\) & 19 & 1 & 11 \\
            \gls{rpc} & 5.23 & \(\bf 8.330 \times 10^{-1}\) & 16 & 0 & 15 \\
            Native-\gls{eps} & 5.53 & \(\bf 4.877 \times 10^{-1}\) & 24 & 0 & 7 \\
            \gls{st}-\gls{pi} & 5.74 & \(\bf 3.220 \times 10^{-1}\) & 21 & 0 & 10 \\
            Chain-\gls{eps} & 6.34 & \(4.773 \times 10^{-2}\) & 23 & 0 & 8 \\
            Native-\gls{pi} & 6.79 & \(7.935 \times 10^{-3}\) & 19 & 1 & 11 \\
            Chain-\gls{pi} & 7.37 & \(4.652 \times 10^{-4}\) & 23 & 0 & 8 \\
            \tabucline[0.1cm]{-}
            \multicolumn{6}{|[0.05cm]c|[0.1cm]}{\textbf{Miss percentage: 30\%}} \\
            \hline
            \multicolumn{6}{|[0.05cm]c|[0.1cm]}{Friedman p-value: \(\bf 1.659 \times 10^{-36}\)} \\
            \hline
            \multicolumn{6}{|[0.05cm]c|[0.1cm]}{Holm results} \\
            \hline
            Method & Rank & P-value & Win & Tie & Loss \\
            \hline
            \gls{st}-\gls{eps} & 2.39 & - & - & - & - \\
            \gls{st}-\gls{rr} & 2.40 & \(\bf 1.000\) & 16 & 1 & 14 \\
            \gls{st}-\gls{pi} & 2.89 & \(\bf 1.000\) & 20 & 1 & 10 \\
            \gls{rpc} & 3.10 & \(\bf 1.000\) & 18 & 0 & 13 \\
            Native-\gls{pi} & 6.65 & \(1.231 \times 10^{-7}\) & 30 & 0 & 1 \\
            Chain-\gls{pi} & 6.74 & \(7.446 \times 10^{-8}\) & 30 & 0 & 1 \\
            Native-\gls{eps} & 6.94 & \(2.051 \times 10^{-8}\) & 31 & 0 & 0 \\
            Chain-\gls{eps} & 6.95 & \(2.051 \times 10^{-8}\) & 31 & 0 & 0 \\
            Native-\gls{rr} & 8.23 & \(2.513 \times 10^{-13}\) & 30 & 0 & 1 \\
            Chain-\gls{rr} & 8.73 & \(1.517 \times 10^{-15}\) & 29 & 0 & 2 \\
            \tabucline[0.1cm]{-}
            \multicolumn{6}{|[0.05cm]c|[0.1cm]}{\textbf{Miss percentage: 60\%}} \\
            \hline
            \multicolumn{6}{|[0.05cm]c|[0.1cm]}{Friedman p-value: \(\bf 2.588 \times 10^{-40}\)} \\
            \hline
            \multicolumn{6}{|[0.05cm]c|[0.1cm]}{Holm results} \\
            \hline
            Method & Rank & P-value & Win & Tie & Loss \\
            \hline
            \gls{st}-\gls{eps} & 2.11 & - & - & - & - \\
            \gls{st}-\gls{rr} & 2.29 & \(\bf 9.516 \times 10^{-1}\) & 18 & 0 & 13 \\
            \gls{st}-\gls{pi} & 2.66 & \(\bf 9.516 \times 10^{-1}\) & 20 & 1 & 10 \\
            \gls{rpc} & 3.29 & \(\bf 3.773 \times 10^{-1}\) & 24 & 0 & 7 \\
            Native-\gls{eps} & 6.21 & \(3.989 \times 10^{-7}\) & 31 & 0 & 0 \\
            Native-\gls{pi} & 6.77 & \(6.751 \times 10^{-9}\) & 30 & 0 & 1 \\
            Chain-\gls{eps} & 7.21 & \(2.047 \times 10^{-10}\) & 31 & 0 & 0 \\
            Chain-\gls{pi} & 7.48 & \(2.006 \times 10^{-11}\) & 30 & 0 & 1 \\
            Native-\gls{rr} & 7.94 & \(2.954 \times 10^{-13}\) & 30 & 0 & 1 \\
            Chain-\gls{rr} & 9.03 & \(2.077 \times 10^{-18}\) & 30 & 0 & 1 \\
            \tabucline[0.1cm]{-}
        \end{tabu}%
    }
    \label{table:tests}
\end{table*}

In light of these results, the \gls{more-plr} estimators perform particularly well in \gls{lr} problems with complete rankings, as they are all ranked ahead of the \gls{rpc} method.
Moreover, with incomplete rankings, the \gls{st}-\gls{pi} method is statistically different from \gls{rpc} and the rest of the \gls{more-plr} learners, except for the \gls{st}-\gls{eps} and \gls{st}-\gls{rr} algorithms, which shows no statistical difference compared to it.
Therefore, for \gls{lr} problems, it can be concluded that the \gls{st} methods provide the best performance.

Looking at the results for the \gls{plr} problems, \gls{rpc} is ranked first with \(0\%\) and \(30\%\) of missing labels but performs worse with \(60\%\), where the \gls{st}-\gls{eps} is ranked first.
Furthermore, with complete rankings, there is no statistical difference between \gls{rpc}, Native-\gls{rr}, Chain-\gls{rr}, and \gls{st}-\gls{rr}.
However, in incomplete rankings, \gls{rpc}, \gls{st}-\gls{rr}, \gls{st}-\gls{eps}, and \gls{st}-\gls{pi} perform equally well.
Therefore, for \gls{plr} problems, these learners are more stable against missing labels than Native and Chain.

Considering both problems simultaneously, the Native-\gls{rr} method ranks first according to the Friedman test in the complete case.
In contrast, the \gls{st}-\gls{eps} is ranked first with \(30\%\) and \(60\%\) of missing labels, with \gls{st}-\gls{rr}, \gls{st}-\gls{pi}, and \gls{rpc} methods not exhibiting any statistical difference.
Overall, the \gls{st} methods show more stability across different problems and varying degrees of missing labels than the rest of the algorithms.

This stability of the \gls{st} methods against incomplete rankings, particularly concerning the \gls{rpc} method, is because when a label is missing for a particular instance in the training dataset, the \gls{st} method only excludes that instance from the learner associated with that specific label.
In contrast, the \gls{rpc} method excludes the instance from all learners considering that label in a pairwise comparison.
For an illustration, take as an example the following data set \(\cD \subseteq \cX \times \mathcal{S}_k^{\succeq, \mathrm{inc}}\) containing two instances \(z_1 = (\x_1, \succeq_{\mathcal{B}_1}) \in \cD\) and \(z_2 = (\x_2, \succeq_{\mathcal{B}_2}) \in \cD\), with associated bucket orders \(\succeq_{\mathcal{B}_1} = \langle \{u_1\}, \{u_2, u_4\} \rangle\) and \(\succeq_{\mathcal{B}_2} = \langle \{u_1\}, \{u_2, u_3\} \rangle\).
The \gls{st} methods will train four models on the following datasets: \(\mathcal{D}_1 = \mathcal{D}_2 = \{z_1, z_2\}\), \(\mathcal{D}_3 = \{z_1\}\), and \(\mathcal{D}_4 = \{z_2\}\).
On the other hand, the \gls{rpc} model will train six models using these datasets\footnote{The subscripts indicate the pairs of items \(u_1, \dots, u_4\) used.}: \(\mathcal{D}_{1,2} = \{z_1, z_2\}\), \(\mathcal{D}_{1,3} = \mathcal{D}_{2,3} = \{z_2\}\), \(\mathcal{D}_{1,4} = \mathcal{D}_{2,4} = \{z_1\}\), and \(\mathcal{D}_{3,4} = \emptyset\).
Thus, \gls{st} trains models on two  datasets containing all training instances, while \gls{rpc} trains models on one dataset.

\subsubsection*{CPU Time}

Table \ref{table:time} provides the time results (in seconds) of the algorithms with complete rankings.

\begin{table*}[t]
    \centering
    \caption{Time results (in seconds) for complete rankings}
    \resizebox{\textwidth}{!}
    {
        \begin{tabu}{lrrrrrrrrrrr}
            \toprule
            Problem & Dataset & \gls{rpc} & \gls{st}-\gls{rr} & \gls{st}-\gls{pi} & \gls{st}-\gls{eps} & Chain-\gls{rr} & Chain-\gls{pi} & Chain-\gls{eps} & Native-\gls{rr} & Native-\gls{pi} & Native-\gls{eps} \\
            \midrule
            \multirow{13}{*}{\gls{lr}} & authorship & 0.637 \(\pm\) 0.141 & 0.825 \(\pm\) 0.137 & 0.872 \(\pm\) 0.299 & 0.915 \(\pm\) 0.147 & 1.420 \(\pm\) 0.008 & 1.396 \(\pm\) 0.009 & 1.416 \(\pm\) 0.007 & 0.537 \(\pm\) 0.014 & 0.532 \(\pm\) 0.017 & 0.534 \(\pm\) 0.013 \\
            & glass & 0.604 \(\pm\) 0.019 & 0.593 \(\pm\) 0.015 & 0.578 \(\pm\) 0.015 & 0.666 \(\pm\) 0.018 & 1.207 \(\pm\) 0.015 & 1.198 \(\pm\) 0.010 & 1.218 \(\pm\) 0.009 & 0.462 \(\pm\) 0.014 & 0.480 \(\pm\) 0.013 & 0.458 \(\pm\) 0.016 \\
            & iris & 0.429 \(\pm\) 0.045 & 0.538 \(\pm\) 0.013 & 0.523 \(\pm\) 0.012 & 0.536 \(\pm\) 0.019 & 0.860 \(\pm\) 0.008 & 0.847 \(\pm\) 0.009 & 0.865 \(\pm\) 0.007 & 0.483 \(\pm\) 0.007 & 0.482 \(\pm\) 0.012 & 0.475 \(\pm\) 0.014 \\
            & letter & 7.808 \(\pm\) 0.042 & 5.364 \(\pm\) 0.068 & 5.203 \(\pm\) 0.022 & 5.490 \(\pm\) 0.040 & 6.503 \(\pm\) 0.019 & 6.552 \(\pm\) 0.019 & 6.625 \(\pm\) 0.034 & 1.413 \(\pm\) 0.013 & 1.262 \(\pm\) 0.023 & 1.282 \(\pm\) 0.011 \\
            & libras & 1.654 \(\pm\) 0.021 & 1.568 \(\pm\) 0.336 & 1.500 \(\pm\) 0.009 & 1.537 \(\pm\) 0.012 & 2.445 \(\pm\) 0.012 & 2.382 \(\pm\) 0.017 & 2.464 \(\pm\) 0.013 & 0.570 \(\pm\) 0.011 & 0.549 \(\pm\) 0.009 & 0.559 \(\pm\) 0.010 \\
            & movies & 1.573 \(\pm\) 0.067 & 0.870 \(\pm\) 0.010 & 0.864 \(\pm\) 0.011 & 0.929 \(\pm\) 0.019 & 2.067 \(\pm\) 0.025 & 2.013 \(\pm\) 0.014 & 2.060 \(\pm\) 0.012 & 0.456 \(\pm\) 0.012 & 0.466 \(\pm\) 0.013 & 0.454 \(\pm\) 0.011 \\
            & pendigits & 2.364 \(\pm\) 0.028 & 2.419 \(\pm\) 0.008 & 2.428 \(\pm\) 0.009 & 2.436 \(\pm\) 0.012 & 2.627 \(\pm\) 0.013 & 2.581 \(\pm\) 0.007 & 2.664 \(\pm\) 0.016 & 1.060 \(\pm\) 0.012 & 1.005 \(\pm\) 0.021 & 1.067 \(\pm\) 0.011 \\
            & political & 0.902 \(\pm\) 0.014 & 0.650 \(\pm\) 0.109 & 0.833 \(\pm\) 0.295 & 0.634 \(\pm\) 0.017 & 1.401 \(\pm\) 0.011 & 1.449 \(\pm\) 0.035 & 1.397 \(\pm\) 0.020 & 0.446 \(\pm\) 0.010 & 0.516 \(\pm\) 0.047 & 0.457 \(\pm\) 0.013 \\
            & segment & 1.066 \(\pm\) 0.020 & 1.172 \(\pm\) 0.007 & 1.153 \(\pm\) 0.007 & 1.195 \(\pm\) 0.011 & 1.435 \(\pm\) 0.013 & 1.378 \(\pm\) 0.010 & 1.455 \(\pm\) 0.009 & 0.601 \(\pm\) 0.011 & 0.579 \(\pm\) 0.013 & 0.593 \(\pm\) 0.006 \\
            & vehicle & 0.553 \(\pm\) 0.020 & 0.627 \(\pm\) 0.012 & 0.604 \(\pm\) 0.010 & 0.680 \(\pm\) 0.029 & 0.959 \(\pm\) 0.010 & 0.932 \(\pm\) 0.012 & 0.950 \(\pm\) 0.009 & 0.461 \(\pm\) 0.010 & 0.462 \(\pm\) 0.016 & 0.463 \(\pm\) 0.006 \\
            & vowel & 1.239 \(\pm\) 0.046 & 0.862 \(\pm\) 0.009 & 0.845 \(\pm\) 0.011 & 0.902 \(\pm\) 0.018 & 1.489 \(\pm\) 0.013 & 1.430 \(\pm\) 0.011 & 1.491 \(\pm\) 0.014 & 0.458 \(\pm\) 0.014 & 0.457 \(\pm\) 0.011 & 0.464 \(\pm\) 0.013 \\
            & wine & 0.432 \(\pm\) 0.038 & 0.511 \(\pm\) 0.013 & 0.511 \(\pm\) 0.015 & 0.548 \(\pm\) 0.025 & 0.864 \(\pm\) 0.011 & 0.846 \(\pm\) 0.011 & 0.870 \(\pm\) 0.008 & 0.469 \(\pm\) 0.010 & 0.481 \(\pm\) 0.011 & 0.462 \(\pm\) 0.009 \\
            & yeast & 1.123 \(\pm\) 0.019 & 0.859 \(\pm\) 0.007 & 0.836 \(\pm\) 0.010 & 0.897 \(\pm\) 0.015 & 1.479 \(\pm\) 0.017 & 1.417 \(\pm\) 0.013 & 1.457 \(\pm\) 0.010 & 0.494 \(\pm\) 0.012 & 0.480 \(\pm\) 0.013 & 0.499 \(\pm\) 0.013 \\
            \midrule
            \multirow{18}{*}{\gls{plr}} & algae & 0.779 \(\pm\) 0.027 & 0.667 \(\pm\) 0.008 & 0.708 \(\pm\) 0.020 & 0.670 \(\pm\) 0.011 & 1.253 \(\pm\) 0.014 & 1.185 \(\pm\) 0.015 & 1.254 \(\pm\) 0.028 & 0.447 \(\pm\) 0.011 & 0.450 \(\pm\) 0.014 & 0.453 \(\pm\) 0.012 \\
            & authorship & 0.652 \(\pm\) 0.111 & 0.852 \(\pm\) 0.127 & 0.968 \(\pm\) 0.249 & 0.839 \(\pm\) 0.117 & 1.459 \(\pm\) 0.008 & 1.457 \(\pm\) 0.010 & 1.545 \(\pm\) 0.036 & 0.551 \(\pm\) 0.010 & 0.560 \(\pm\) 0.019 & 0.577 \(\pm\) 0.036 \\
            & blocks & 1.357 \(\pm\) 0.025 & 1.371 \(\pm\) 0.009 & 1.189 \(\pm\) 0.021 & 1.164 \(\pm\) 0.010 & 1.315 \(\pm\) 0.009 & 1.339 \(\pm\) 0.010 & 1.525 \(\pm\) 0.055 & 0.751 \(\pm\) 0.012 & 0.768 \(\pm\) 0.016 & 0.814 \(\pm\) 0.032 \\
            & breast & 0.596 \(\pm\) 0.016 & 0.587 \(\pm\) 0.011 & 0.598 \(\pm\) 0.019 & 0.573 \(\pm\) 0.009 & 1.203 \(\pm\) 0.013 & 1.128 \(\pm\) 0.026 & 1.137 \(\pm\) 0.036 & 0.475 \(\pm\) 0.011 & 0.472 \(\pm\) 0.012 & 0.432 \(\pm\) 0.012 \\
            & ecoli & 0.808 \(\pm\) 0.019 & 0.668 \(\pm\) 0.013 & 0.675 \(\pm\) 0.018 & 0.643 \(\pm\) 0.015 & 1.384 \(\pm\) 0.011 & 1.296 \(\pm\) 0.029 & 1.339 \(\pm\) 0.033 & 0.476 \(\pm\) 0.014 & 0.466 \(\pm\) 0.009 & 0.440 \(\pm\) 0.013 \\
            & glass & 0.605 \(\pm\) 0.015 & 0.612 \(\pm\) 0.016 & 0.613 \(\pm\) 0.016 & 0.590 \(\pm\) 0.010 & 1.194 \(\pm\) 0.011 & 1.114 \(\pm\) 0.029 & 1.144 \(\pm\) 0.027 & 0.469 \(\pm\) 0.013 & 0.459 \(\pm\) 0.013 & 0.431 \(\pm\) 0.018 \\
            & iris & 0.401 \(\pm\) 0.011 & 0.541 \(\pm\) 0.014 & 0.518 \(\pm\) 0.027 & 0.542 \(\pm\) 0.017 & 0.858 \(\pm\) 0.009 & 0.807 \(\pm\) 0.016 & 0.771 \(\pm\) 0.072 & 0.490 \(\pm\) 0.009 & 0.480 \(\pm\) 0.010 & 0.451 \(\pm\) 0.011 \\
            & letter & 10.377 \(\pm\) 0.713 & 5.737 \(\pm\) 0.047 & 5.162 \(\pm\) 0.059 & 4.954 \(\pm\) 0.037 & 5.955 \(\pm\) 0.015 & 6.122 \(\pm\) 0.014 & 6.719 \(\pm\) 0.266 & 1.247 \(\pm\) 0.007 & 1.240 \(\pm\) 0.027 & 1.311 \(\pm\) 0.025 \\
            & libras & 1.844 \(\pm\) 0.138 & 1.584 \(\pm\) 0.006 & 1.581 \(\pm\) 0.011 & 1.567 \(\pm\) 0.006 & 2.481 \(\pm\) 0.009 & 2.458 \(\pm\) 0.012 & 2.698 \(\pm\) 0.067 & 0.553 \(\pm\) 0.009 & 0.570 \(\pm\) 0.014 & 0.587 \(\pm\) 0.037 \\
            & movies & 1.562 \(\pm\) 0.028 & 0.838 \(\pm\) 0.010 & 0.898 \(\pm\) 0.016 & 0.877 \(\pm\) 0.011 & 2.083 \(\pm\) 0.014 & 1.961 \(\pm\) 0.023 & 2.078 \(\pm\) 0.072 & 0.456 \(\pm\) 0.013 & 0.451 \(\pm\) 0.010 & 0.459 \(\pm\) 0.013 \\
            & pendigits & 3.332 \(\pm\) 0.043 & 2.796 \(\pm\) 0.020 & 2.576 \(\pm\) 0.015 & 2.545 \(\pm\) 0.010 & 2.782 \(\pm\) 0.008 & 2.768 \(\pm\) 0.005 & 3.111 \(\pm\) 0.112 & 1.042 \(\pm\) 0.013 & 1.051 \(\pm\) 0.014 & 1.114 \(\pm\) 0.053 \\
            & political & 1.068 \(\pm\) 0.023 & 0.969 \(\pm\) 0.091 & 1.191 \(\pm\) 0.379 & 0.916 \(\pm\) 0.009 & 1.557 \(\pm\) 0.011 & 1.924 \(\pm\) 0.368 & 1.563 \(\pm\) 0.031 & 0.560 \(\pm\) 0.017 & 0.690 \(\pm\) 0.220 & 0.572 \(\pm\) 0.023 \\
            & satimage & 1.658 \(\pm\) 0.013 & 2.078 \(\pm\) 0.008 & 2.047 \(\pm\) 0.012 & 2.022 \(\pm\) 0.009 & 2.273 \(\pm\) 0.008 & 2.289 \(\pm\) 0.007 & 2.521 \(\pm\) 0.084 & 1.002 \(\pm\) 0.010 & 1.005 \(\pm\) 0.013 & 1.054 \(\pm\) 0.053 \\
            & segment & 1.282 \(\pm\) 0.019 & 1.255 \(\pm\) 0.007 & 1.148 \(\pm\) 0.016 & 1.131 \(\pm\) 0.008 & 1.466 \(\pm\) 0.010 & 1.417 \(\pm\) 0.011 & 1.552 \(\pm\) 0.054 & 0.608 \(\pm\) 0.008 & 0.621 \(\pm\) 0.016 & 0.636 \(\pm\) 0.034 \\
            & vehicle & 0.620 \(\pm\) 0.031 & 0.653 \(\pm\) 0.009 & 0.669 \(\pm\) 0.036 & 0.642 \(\pm\) 0.007 & 0.962 \(\pm\) 0.010 & 0.918 \(\pm\) 0.012 & 1.036 \(\pm\) 0.087 & 0.467 \(\pm\) 0.011 & 0.465 \(\pm\) 0.006 & 0.492 \(\pm\) 0.026 \\
            & vowel & 1.269 \(\pm\) 0.015 & 0.872 \(\pm\) 0.009 & 0.908 \(\pm\) 0.015 & 0.886 \(\pm\) 0.012 & 1.504 \(\pm\) 0.013 & 1.439 \(\pm\) 0.012 & 1.630 \(\pm\) 0.144 & 0.459 \(\pm\) 0.009 & 0.462 \(\pm\) 0.016 & 0.478 \(\pm\) 0.014 \\
            & wine & 0.434 \(\pm\) 0.048 & 0.516 \(\pm\) 0.012 & 0.534 \(\pm\) 0.032 & 0.517 \(\pm\) 0.010 & 0.858 \(\pm\) 0.013 & 0.813 \(\pm\) 0.020 & 0.788 \(\pm\) 0.021 & 0.474 \(\pm\) 0.012 & 0.461 \(\pm\) 0.011 & 0.460 \(\pm\) 0.028 \\
            & yeast & 1.212 \(\pm\) 0.022 & 0.921 \(\pm\) 0.010 & 0.933 \(\pm\) 0.023 & 0.903 \(\pm\) 0.007 & 1.491 \(\pm\) 0.012 & 1.419 \(\pm\) 0.011 & 1.553 \(\pm\) 0.071 & 0.486 \(\pm\) 0.010 & 0.483 \(\pm\) 0.011 & 0.501 \(\pm\) 0.014 \\
            \bottomrule
        \end{tabu}
    \label{table:time}
    }
\end{table*}
These results conclude that the fastest algorithms are the Native variants, as they only fit one model.
Additionally, the results show that the \gls{st} techniques are faster overall concerning \gls{rpc}, particularly with datasets such as letter.
This is expected, as \(k\) models need to be fitted for \gls{st} in comparison to \(\binom{k}{2}\) in \gls{rpc}.
The \gls{rpc} approach sometimes demonstrates faster performance on datasets with a small number of labels (see iris or authorship), as \gls{st} and \gls{rpc} fit a similar number of models in these cases.
The difference in complexity can be attributed to the \gls{plr} post-hoc layers, which act as minor bottlenecks, resulting in complexities of \(\bO(k)\) for \gls{rr} and \(\bO(k^2)\) for \gls{pi} and \gls{eps}.
Importantly, the transformation step of \gls{rpc} requires solving the \gls{obop} problem, which has a complexity of \(\bO(klog(k))\).

Given that the random forests algorithm has a complexity of \(\bO(bmnlog(n))\), where \(b\) represents the number of decision trees, \(m\) the number of features, and \(n\) the number of instances, the complexities are as follows.
\begin{itemize}
    \item \gls{rpc}: \(\bO(k^2bmnlog(n) + klog(k))\),
    \item \gls{st}-\gls{rr} and Chain-\gls{rr}: \(\bO(kbmnlog(n) + k)\),
    \item \gls{st}-\gls{pi}, \gls{st}-\gls{eps}, Chain-\gls{pi}, and Chain-\gls{eps}: \(\bO(kbmnlog(n) + k^2)\),
    \item Native-\gls{rr}: \(\bO(bmnlog(n) + k)\),
    \item Native-\gls{pi} and Native-\gls{eps}: \(\bO(bmnlog(n) + k^2)\).
\end{itemize}

Note that these theoretical computational complexities align with the time results, confirming the expected patterns.
The same conclusions hold for incomplete rankings, with lower times due to fewer instances used for training the models.

        \section{Conclusions}
        \label{section:conclusions}
        This paper has explored using \gls{mor} for the \gls{plr} problem.
Given the multiple approaches available, we have designed a framework specifically for this purpose and detailed its key components. 
In addition to the initial version of the framework, we have included a new component that acts as an encoder for transforming \gls{plr} datasets into \gls{mor} datasets.
Our experiments demonstrate that, with the right combination of components, the framework can outperform state-of-the-art \gls{plr} methods, achieving better results in less time.

A promising direction for future research would be the automated selection of the optimal configuration and hyperparameters for the proposed \gls{more-plr} framework using the meta-information of a dataset.
This could potentially be accomplished using automated machine learning techniques \parencite{karl_automated_2024}.
Additionally, as discussed in the comparison between our framework and traditional \gls{plr} learners, the encoder in the latter is fixed.
Exploring alternative encoding options within this context would be valuable.
Another promising direction would be the development of a hybrid learner that combines the strengths of both frameworks, leveraging the benefits of each.
Finally, applying the investigated methods in real-world scenarios to aid human decision-making presents a compelling avenue for future work. One such example is using these methods to rank prompts when fine-tuning \glspl{llm}.

    \backmatter

    \section*{Declarations}

    \bmhead{Funding}

    This work is partially funded by the following projects: SBPLY/21/180225/000062 (Junta de Comunidades de Castilla-La Mancha and ERDF A way of making Europe), PID2022-139293NB-C32 (MICIU/AEI/10.13039/501100011033 and ERDF, EU), and 2022-GRIN-34437 (Universidad de Castilla-La Mancha and ERDF A way of making Europe).

    \bmhead{Author contributions}

    Conceptualization, S.M.A.R.T., J.C.A., and V.B.; Data curation, S.M.A.R.T. and J.C.A.; Formal analysis, S.M.A.R.T., J.C.A., and  V.B.; Investigation, S.M.A.R.T., J.C.A., and V.B.; Methodology, S.M.A.R.T., J.C.A., and V.B.; Software, S.M.A.R.T. and J.C.A.; Supervision, J.C.A and V.B.; Validation, S.M.A.R.T. and J.C.A.; Visualization, S.M.A.R.T., J.C.A., and V.B.; Writing—original draft, S.M.A.R.T., J.C.A., and V.B.; Writing—review and editing, S.M.A.R.T., J.C.A., and V.B.

    \bmhead{Data availability}

    The datasets are publicly available at \url{https://www.openml.org}.

    \bmhead{Code availability}

    The code is available open-access at \url{https://github.com/Advueu963/MORE-PLR}.
    
    \bmhead{Conflict of interest}
    
    The authors declare no conflict of interest.
    
    \printbibliography

    \clearpage
    \newpage

    \appendix

        \section{Transforming Buckets Orders to Rank Position Vectors}
        \label{appendix:encoding}
        Using the bucket order \(\B = \langle\{u_1\}, \{u_2, u_3\}, \{u_4\}\rangle\), we compute the corresponding rank position vector \(\vec{p} = (p_1, \dots, p_4)\) for each of the encoding methods as detailed below.

\paragraph{Dense}
    \begin{equation*}
        p_1 = 1 + \underbrace{\id_{\{\bucket{1} \succ_{\buckets} u_1\}}}_{0} 
        + \underbrace{\id_{\{\bucket{2} \succ_{\buckets} u_1\}}}_{0} 
        + \underbrace{\id_{\{\bucket{3} \succ_{\buckets} u_1\}}}_{0} 
        = 1
    \end{equation*}
    \begin{equation*}
        p_2 = 1 + \underbrace{\id_{\{\bucket{1} \succ_{\buckets} u_2\}}}_{1} 
        + \underbrace{\id_{\{\bucket{2} \succ_{\buckets} u_2\}}}_{0} 
        + \underbrace{\id_{\{\bucket{3} \succ_{\buckets} u_2\}}}_{0} 
        = 2
    \end{equation*}
    \begin{equation*}
        p_3 = 1 + \underbrace{\id_{\{\bucket{1} \succ_{\buckets} u_3\}}}_{1} 
        + \underbrace{\id_{\{\bucket{2} \succ_{\buckets} u_3\}}}_{0} 
        + \underbrace{\id_{\{\bucket{3} \succ_{\buckets} u_3\}}}_{0} 
        = 2%
    \end{equation*}
    \begin{equation*}
        p_4 = 1 + \underbrace{\id_{\{\bucket{1} \succ_{\buckets} u_4\}}}_{1} 
        + \underbrace{\id_{\{\bucket{2} \succ_{\buckets} u_4\}}}_{1} 
        + \underbrace{\id_{\{\bucket{3} \succ_{\buckets} u_4\}}}_{0} 
        = 3
    \end{equation*}

\paragraph{Standard competition}
    \begin{equation*}
        p_1 = 1 
            + \underbrace{\underbrace{\id_{\{\bucket{1} \succ_{\buckets} u_1\}}}_{0} \cdot \underbrace{\left|\bucket{1}\right|}_{1}}_{0} 
            + \underbrace{\underbrace{\id_{\{\bucket{2} \succ_{\buckets} u_1\}}}_{0} \cdot \underbrace{\left|\bucket{2}\right|}_{2}}_{0} 
            + \underbrace{\underbrace{\id_{\{\bucket{3} \succ_{\buckets} u_1\}}}_{0} \cdot \underbrace{\left|\bucket{3}\right|}_{1}}_{0} 
            = 1
    \end{equation*}
    \begin{equation*}
        p_2 = 1 
            + \underbrace{\underbrace{\id_{\{\bucket{1} \succ_{\buckets} u_2\}}}_{1} \cdot \underbrace{\left|\bucket{1}\right|}_{1}}_{1} 
            + \underbrace{\underbrace{\id_{\{\bucket{2} \succ_{\buckets} u_2\}}}_{0} \cdot \underbrace{\left|\bucket{2}\right|}_{2}}_{0} 
            + \underbrace{\underbrace{\id_{\{\bucket{3} \succ_{\buckets} u_2\}}}_{0} \cdot \underbrace{\left|\bucket{3}\right|}_{1}}_{0} 
            = 2
    \end{equation*}
    \begin{equation*}
        p_3 = 1 
            + \underbrace{\underbrace{\id_{\{\bucket{1} \succ_{\buckets} u_3\}}}_{1} \cdot \underbrace{\left|\bucket{1}\right|}_{1}}_{1} 
            + \underbrace{\underbrace{\id_{\{\bucket{2} \succ_{\buckets} u_3\}}}_{0} \cdot \underbrace{\left|\bucket{2}\right|}_{2}}_{0} 
            + \underbrace{\underbrace{\id_{\{\bucket{3} \succ_{\buckets} u_3\}}}_{0} \cdot \underbrace{\left|\bucket{3}\right|}_{1}}_{0} 
            = 2
    \end{equation*}
    \begin{equation*}
        p_4 = 1 
        + \underbrace{\underbrace{\id_{\{\bucket{1} \succ_{\buckets} u_4\}}}_{1} \cdot \underbrace{\left|\bucket{1}\right|}_{1}}_{1} 
        + \underbrace{\underbrace{\id_{\{\bucket{2} \succ_{\buckets} u_4\}}}_{1} \cdot \underbrace{\left|\bucket{2}\right|}_{2}}_{2} 
        + \underbrace{\underbrace{\id_{\{\bucket{3} \succ_{\buckets} u_4\}}}_{0} \cdot \underbrace{\left|\bucket{3}\right|}_{1}}_{0} 
        = 4
    \end{equation*}

\paragraph{Modified competition}
    \begin{equation*}
        \resizebox{\textwidth}{!}
        {
            \(
                p_1 = \underbrace{\underbrace{\id_{\{\bucket{1} \succ_{\buckets} u_1\}}}_{0} \cdot \underbrace{|\bucket{1}|}_{1}}_{0} 
                + \underbrace{\underbrace{\id_{\{u_1 \in \buckets_1\}}}_{1} \cdot \underbrace{|\bucket{1}|}_{1}}_{1}
                + \underbrace{\underbrace{\id_{\{\bucket{2} \succ_{\buckets} u_1\}}}_{0} \cdot \underbrace{|\bucket{2}|}_{2}}_{0} 
                + \underbrace{\underbrace{\id_{\{u_1 \in \buckets_2\}}}_{0} \cdot \underbrace{|\bucket{2}|}_{2}}_{0}
                + \underbrace{\underbrace{\id_{\{\bucket{3} \succ_{\buckets} u_1\}}}_{0} \cdot \underbrace{|\bucket{3}|}_{1}}_{0} 
                + \underbrace{\underbrace{\id_{\{u_1 \in \buckets_3\}}}_{0} \cdot \underbrace{|\bucket{3}|}_{1}}_{0} = 1
            \)
        }
    \end{equation*}
    \begin{equation*}
        \resizebox{\textwidth}{!}{
            \(
                p_2 = \underbrace{\underbrace{\id_{\{\bucket{1} \succ_{\buckets} u_2\}}}_{1} \cdot \underbrace{|\bucket{1}|}_{1}}_{1} 
                + \underbrace{\underbrace{\id_{\{u_2 \in \buckets_1\}}}_{0} \cdot \underbrace{|\bucket{1}|}_{1}}_{0}
                + \underbrace{\underbrace{\id_{\{\bucket{2} \succ_{\buckets} u_2\}}}_{0} \cdot \underbrace{|\bucket{2}|}_{2}}_{0} 
                + \underbrace{\underbrace{\id_{\{u_2 \in \buckets_2\}}}_{1} \cdot \underbrace{|\bucket{2}|}_{2}}_{2}
                + \underbrace{\underbrace{\id_{\{\bucket{3} \succ_{\buckets} u_2\}}}_{0} \cdot \underbrace{|\bucket{3}|}_{1}}_{0} 
                + \underbrace{\underbrace{\id_{\{u_2 \in \buckets_3\}}}_{0} \cdot \underbrace{|\bucket{3}|}_{1}}_{0} = 3
            \)
        }
    \end{equation*}
    \begin{equation*}
        \resizebox{\textwidth}{!}{
            \(
                p_3 = \underbrace{\underbrace{\id_{\{\bucket{1} \succ_{\buckets} u_3\}}}_{1} \cdot \underbrace{|\bucket{1}|}_{1}}_{1} 
                + \underbrace{\underbrace{\id_{\{u_3 \in \buckets_1\}}}_{0} \cdot \underbrace{|\bucket{1}|}_{1}}_{0}
                + \underbrace{\underbrace{\id_{\{\bucket{2} \succ_{\buckets} u_3\}}}_{0} \cdot \underbrace{|\bucket{2}|}_{2}}_{0} 
                + \underbrace{\underbrace{\id_{\{u_3 \in \buckets_2\}}}_{1} \cdot \underbrace{|\bucket{2}|}_{2}}_{2}
                + \underbrace{\underbrace{\id_{\{\bucket{3} \succ_{\buckets} u_3\}}}_{0} \cdot \underbrace{|\bucket{3}|}_{1}}_{0} 
                + \underbrace{\underbrace{\id_{\{u_3 \in \buckets_3\}}}_{0} \cdot \underbrace{|\bucket{3}|}_{1}}_{0} = 3
            \)
        }
    \end{equation*}
    \begin{equation*}
        \resizebox{\textwidth}{!}{
            \(
                p_4 = \underbrace{\underbrace{\id_{\{\bucket{1} \succ_{\buckets} u_4\}}}_{1} \cdot \underbrace{|\bucket{1}|}_{1}}_{1} 
                + \underbrace{\underbrace{\id_{\{u_4 \in \buckets_1\}}}_{0} \cdot \underbrace{|\bucket{1}|}_{1}}_{0}
                + \underbrace{\underbrace{\id_{\{\bucket{2} \succ_{\buckets} u_4\}}}_{1} \cdot \underbrace{|\bucket{2}|}_{2}}_{2} 
                + \underbrace{\underbrace{\id_{\{u_4 \in \buckets_2\}}}_{0} \cdot \underbrace{|\bucket{2}|}_{2}}_{0}
                + \underbrace{\underbrace{\id_{\{\bucket{3} \succ_{\buckets} u_4\}}}_{0} \cdot \underbrace{|\bucket{3}|}_{1}}_{0} 
                + \underbrace{\underbrace{\id_{\{u_4 \in \buckets_3\}}}_{1} \cdot \underbrace{|\bucket{3}|}_{1}}_{1} = 4
            \)
        }
    \end{equation*}

\paragraph{Fractional}
    \noindent%
    \resizebox{0.5\textwidth}{!}
    {%
        $
        \displaystyle
        \begin{aligned}
            p_1 = &\underbrace{\underbrace{\id_{\{\bucket{1} \succ_{\buckets} u_1\}}}_{0} \cdot \underbrace{|\bucket{1}|}_{1}}_{0} 
            + \underbrace{\underbrace{\id_{\{u_1 \in \buckets_1\}}}_{1} \cdot \underbrace{(1 + (|\bucket{1}| - 1) \cdot 0.5)}_{1}}_{1} \\
            &+ \underbrace{\underbrace{\id_{\{\bucket{2} \succ_{\buckets} u_1\}}}_{0} \cdot \underbrace{|\bucket{2}|}_{2}}_{0} 
            + \underbrace{\underbrace{\id_{\{u_1 \in \buckets_2\}}}_{0} \cdot \underbrace{(1 + (|\bucket{2}| - 1) \cdot 0.5)}_{1.5}}_{0} \\
            &+ \underbrace{\underbrace{\id_{\{\bucket{3} \succ_{\buckets} u_1\}}}_{0} \cdot \underbrace{|\bucket{3}|}_{1}}_{0} 
            + \underbrace{\underbrace{\id_{\{u_1 \in \buckets_3\}}}_{0} \cdot \underbrace{(1 + (|\bucket{3}| - 1) \cdot 0.5)}_{1}}_{0} = 1
        \end{aligned}
        $%
    }%
    \hfill%
    \resizebox{0.5\textwidth}{!}
    {%
        $\displaystyle
        \begin{aligned}
            p_2 = &\underbrace{\underbrace{\id_{\{\bucket{1} \succ_{\buckets} u_2\}}}_{1} \cdot \underbrace{|\bucket{1}|}_{1}}_{1} 
            + \underbrace{\underbrace{\id_{\{u_2 \in \buckets_1\}}}_{0} \cdot \underbrace{(1 + (|\bucket{1}| - 1) \cdot 0.5)}_{1}}_{0} \\
            &+ \underbrace{\underbrace{\id_{\{\bucket{2} \succ_{\buckets} u_2\}}}_{0} \cdot \underbrace{|\bucket{2}|}_{2}}_{0} 
            + \underbrace{\underbrace{\id_{\{u_2 \in \buckets_2\}}}_{1} \cdot \underbrace{(1 + (|\bucket{2}| - 1) \cdot 0.5)}_{1.5}}_{1.5} \\
            &+ \underbrace{\underbrace{\id_{\{\bucket{3} \succ_{\buckets} u_2\}}}_{0} \cdot \underbrace{|\bucket{3}|}_{1}}_{0} 
            + \underbrace{\underbrace{\id_{\{u_2 \in \buckets_3\}}}_{0} \cdot \underbrace{(1 + (|\bucket{3}| - 1) \cdot 0.5)}_{1}}_{0} = 2.5
        \end{aligned}
        $%
    }
    \vspace{1em}
    \noindent%
    \resizebox{0.5\textwidth}{!}
    {%
        $\displaystyle
        \begin{aligned}
            p_3 = &\underbrace{\underbrace{\id_{\{\bucket{1} \succ_{\buckets} u_3\}}}_{1} \cdot \underbrace{|\bucket{1}|}_{1}}_{1} 
            + \underbrace{\underbrace{\id_{\{u_3 \in \buckets_1\}}}_{0} \cdot \underbrace{(1 + (|\bucket{1}| - 1) \cdot 0.5)}_{1}}_{0} \\
            &+ \underbrace{\underbrace{\id_{\{\bucket{2} \succ_{\buckets} u_3\}}}_{0} \cdot \underbrace{|\bucket{2}|}_{2}}_{0} 
            + \underbrace{\underbrace{\id_{\{u_3 \in \buckets_2\}}}_{1} \cdot \underbrace{(1 + (|\bucket{2}| - 1) \cdot 0.5)}_{1.5}}_{1.5} \\
            &+ \underbrace{\underbrace{\id_{\{\bucket{3} \succ_{\buckets} u_3\}}}_{0} \cdot \underbrace{|\bucket{3}|}_{1}}_{0} 
            + \underbrace{\underbrace{\id_{\{u_3 \in \buckets_3\}}}_{0} \cdot \underbrace{(1 + (|\bucket{3}| - 1) \cdot 0.5)}_{1}}_{0} = 2.5
        \end{aligned}
        $%
    }
    \hfill%
    \resizebox{0.5\textwidth}{!}
    {%
        $\displaystyle
        \begin{aligned}
            p_4 = &\underbrace{\underbrace{\id_{\{\bucket{1} \succ_{\buckets} u_4\}}}_{1} \cdot \underbrace{|\bucket{1}|}_{1}}_{1} 
             + \underbrace{\underbrace{\id_{\{u_4 \in \buckets_1\}}}_{0} \cdot \underbrace{(1 + (|\bucket{1}| - 1) \cdot 0.5)}_{1}}_{0} \\
             &+ \underbrace{\underbrace{\id_{\{\bucket{2} \succ_{\buckets} u_4\}}}_{1} \cdot \underbrace{|\bucket{2}|}_{2}}_{2} 
             + \underbrace{\underbrace{\id_{\{u_4 \in \buckets_2\}}}_{0} \cdot \underbrace{(1 + (|\bucket{2}| - 1) \cdot 0.5)}_{1.5}}_{0} \\
             &+ \underbrace{\underbrace{\id_{\{\bucket{3} \succ_{\buckets} u_4\}}}_{0} \cdot \underbrace{|\bucket{3}|}_{1}}_{0} 
             + \underbrace{\underbrace{\id_{\{u_4 \in \buckets_3\}}}_{1} \cdot \underbrace{(1 + (|\bucket{3}| - 1) \cdot 0.5)}_{1}}_{1} = 4
        \end{aligned}
        $%
    }

        \newpage

        \section{Datasets Description}
        \label{appendix:datasets}
        Table \ref{table:datasets} shows the description of the datasets used as \textit{benchmark}. The columns display the following information: Identifier; unique identifier from the \texttt{OpenML} repository \parencite{vanschoren_openml_2014}, \#of instances; number of instances; \#of features, number of input variables describing each instance; \#of labels, number of labels; Unique \#of rankings, number of distinct (partial) label rankings; and Mean \#of buckets, average number of buckets within each ranking.

\begin{table*}[!ht]
    \centering
    \caption{Datasets used for the experiments. The algae, movies, and political datasets correspond to real-world problems.}
    \resizebox{\textwidth}{!}
    {%
        \begin{tabu}{llrrrrrr}
            \toprule
            Problem & Name & Identifier & \#of instances & \#of features & \#of labels & \#of rankings & \(\overline{\mbox{\#of buckets}}\) \\
            \midrule
            \multirow{12}{*}{\gls{lr}} & authorship & 42834 & 841 & 70 & 4 & 17 & 4.0 \\ 
            & glass & 42847 & 214 & 9 & 6 & 30 & 6.0 \\ 
            & iris  & 42851 & 150 & 4 & 3 & 5 & 3.0 \\ 
            & letter & 45727 & 20000 & 16 & 26 & 17806 & 26.0 \\ 
            & libras & 45736 & 360 & 90 & 15 & 317 & 15.0 \\ 
            & pendigits & 42856 & 10992 & 16 & 10 & 2081 & 10.0 \\ 
            & segment & 42859 & 2310 & 18 & 7 & 135 & 7.0 \\ 
            & vehicle & 42863 & 846 & 18 & 4 & 18 & 4.0 \\ 
            & vowel & 42865 & 528 & 10 & 11 & 294 & 11.0 \\ 
            & wine & 42867 & 178 & 13 & 3 & 5 & 3.0 \\ 
            & yeast & 45737 & 1484 & 8 & 10 & 471 & 10.0 \\ 
            & movies & 45735 & 260 & 64 & 15 & 256 & 15.0 \\ 
            & political & \(\na\) & 170 & 46 & 6 & 87 & 6.0 \\
            \midrule
            \multirow{18}{*}{\gls{plr}} & authorship & 42835 & 841 & 70 & 4 & 47 & 3.063 \\
            & blocks & 42836 & 5472 & 10 & 5 & 116 & 2.337 \\ 
            & breast & 42838 & 106 & 9 & 6 & 62 & 3.925 \\ 
            & ecoli & 42844 & 336 & 7 & 8 & 179 & 4.14 \\ 
            & glass & 42848 & 214 & 9 & 6 & 105 & 4.089 \\ 
            & iris & 42871 & 150 & 4 & 3 & 7 & 2.38 \\ 
            & letter & 42853 & 20000 & 16 & 26 & 15014 & 7.033 \\ 
            & libras & 42855 & 360 & 90 & 15 & 356 & 6.889 \\ 
            & pendigits & 42857 & 10992 & 16 & 10 & 3327 & 3.397 \\ 
            & satimage & 42858 & 6435 & 36 & 6 & 504 & 3.356 \\ 
            & segment & 42860 & 2310 & 18 & 7 & 271 & 3.031 \\ 
            & vehicle & 42864 & 846 & 18 & 4 & 47 & 3.117 \\ 
            & vowel & 42866 & 528 & 10 & 11 & 504 & 5.739 \\ 
            & wine & 42872 & 178 & 13 & 3 & 11 & 2.68 \\ 
            & yeast & 42870 & 1484 & 8 & 10 & 1006 & 5.929 \\ 
            & algae & 45755 & 316 & 18 & 7 & 228 & 4.877 \\ 
            & movies & 45738 & 260 & 64 & 15 & 257 & 3.046 \\ 
            & political & \(\na\) & 1053 & 46 & 6 & 364 & 3.787 \\
            \bottomrule
        \end{tabu}
        \label{table:datasets}
    }%
\end{table*}

        \newpage

        \section{Complete Accuracy Results}
        \label{appendix:results}
        Tables \ref{table:accuracy_0}, \ref{table:accuracy_30}, and \ref{table:accuracy_60} present the mean accuracy and standard deviation, averaged over test datasets from repeated cross-validation, with the best score(s) per dataset in bold.

\begin{table*}[!ht]
    \centering
    \caption{Accuracy results for complete rankings}
    \resizebox{\textwidth}{!}
    {
        \begin{tabu}{lrrrrrrrrrrr}
            \toprule
            Problem & Dataset & \gls{rpc} & \gls{st}-\gls{rr} & \gls{st}-\gls{pi} & \gls{st}-\gls{eps} & Chain-\gls{rr} & Chain-\gls{pi} & Chain-\gls{eps} & Native-\gls{rr} & Native-\gls{pi} & Native-\gls{eps} \\
            \midrule
            \multirow{13}{*}{\gls{lr}} & authorship & 0.887 \(\pm\) 0.019 & 0.917 \(\pm\) 0.018 & \bf 0.925 \(\pm\) 0.019 & 0.923 \(\pm\) 0.019 & 0.916 \(\pm\) 0.020 & 0.919 \(\pm\) 0.019 & 0.919 \(\pm\) 0.020 & 0.914 \(\pm\) 0.019 & 0.918 \(\pm\) 0.020 & 0.917 \(\pm\) 0.020 \\
            & glass & 0.856 \(\pm\) 0.047 & 0.893 \(\pm\) 0.036 & 0.901 \(\pm\) 0.037 & 0.898 \(\pm\) 0.036 & 0.883 \(\pm\) 0.041 & 0.890 \(\pm\) 0.041 & 0.889 \(\pm\) 0.042 & 0.902 \(\pm\) 0.032 & \bf 0.907 \(\pm\) 0.031 & 0.904 \(\pm\) 0.030 \\
            & iris & 0.948 \(\pm\) 0.037 & 0.962 \(\pm\) 0.033 & 0.965 \(\pm\) 0.031 & 0.965 \(\pm\) 0.032 & 0.963 \(\pm\) 0.034 & 0.962 \(\pm\) 0.036 & 0.962 \(\pm\) 0.036 & 0.968 \(\pm\) 0.032 & \bf 0.969 \(\pm\) 0.030 & \bf 0.969 \(\pm\) 0.030 \\
            & letter & 0.902 \(\pm\) 0.002 & 0.940 \(\pm\) 0.001 & \bf 0.942 \(\pm\) 0.001 & 0.703 \(\pm\) 0.005 & 0.932 \(\pm\) 0.001 & 0.934 \(\pm\) 0.001 & 0.675 \(\pm\) 0.005 & 0.919 \(\pm\) 0.001 & 0.919 \(\pm\) 0.002 & 0.686 \(\pm\) 0.005 \\
            & libras & 0.856 \(\pm\) 0.017 & 0.898 \(\pm\) 0.013 & \bf 0.905 \(\pm\) 0.012 & 0.874 \(\pm\) 0.012 & 0.898 \(\pm\) 0.014 & 0.904 \(\pm\) 0.014 & 0.876 \(\pm\) 0.014 & 0.883 \(\pm\) 0.016 & 0.889 \(\pm\) 0.015 & 0.861 \(\pm\) 0.014 \\
            & movies & 0.218 \(\pm\) 0.027 & 0.359 \(\pm\) 0.033 & 0.350 \(\pm\) 0.032 & 0.354 \(\pm\) 0.033 & 0.364 \(\pm\) 0.042 & 0.359 \(\pm\) 0.041 & 0.359 \(\pm\) 0.040 & \bf 0.370 \(\pm\) 0.033 & 0.358 \(\pm\) 0.033 & 0.366 \(\pm\) 0.033 \\
            & pendigits & 0.958 \(\pm\) 0.001 & 0.971 \(\pm\) 0.001 & \bf 0.976 \(\pm\) 0.001 & 0.966 \(\pm\) 0.001 & 0.969 \(\pm\) 0.001 & 0.971 \(\pm\) 0.001 & 0.969 \(\pm\) 0.001 & 0.963 \(\pm\) 0.002 & 0.966 \(\pm\) 0.001 & 0.957 \(\pm\) 0.001 \\
            & political & 0.532 \(\pm\) 0.054 & 0.679 \(\pm\) 0.050 & 0.703 \(\pm\) 0.058 & \bf 0.708 \(\pm\) 0.056 & 0.666 \(\pm\) 0.056 & 0.689 \(\pm\) 0.061 & 0.687 \(\pm\) 0.063 & 0.648 \(\pm\) 0.057 & 0.674 \(\pm\) 0.056 & 0.680 \(\pm\) 0.057 \\
            & segment & 0.963 \(\pm\) 0.004 & 0.972 \(\pm\) 0.003 & \bf 0.976 \(\pm\) 0.003 & 0.973 \(\pm\) 0.003 & 0.970 \(\pm\) 0.004 & 0.971 \(\pm\) 0.004 & 0.971 \(\pm\) 0.004 & 0.972 \(\pm\) 0.004 & 0.974 \(\pm\) 0.004 & 0.971 \(\pm\) 0.004 \\
            & vehicle & 0.833 \(\pm\) 0.022 & 0.864 \(\pm\) 0.021 & 0.885 \(\pm\) 0.019 & \bf 0.886 \(\pm\) 0.019 & 0.865 \(\pm\) 0.028 & 0.871 \(\pm\) 0.028 & 0.871 \(\pm\) 0.028 & 0.861 \(\pm\) 0.023 & 0.879 \(\pm\) 0.023 & 0.879 \(\pm\) 0.024 \\
            & vowel & 0.811 \(\pm\) 0.021 & 0.877 \(\pm\) 0.014 & \bf 0.887 \(\pm\) 0.014 & 0.874 \(\pm\) 0.014 & 0.856 \(\pm\) 0.018 & 0.861 \(\pm\) 0.018 & 0.852 \(\pm\) 0.018 & 0.870 \(\pm\) 0.017 & 0.877 \(\pm\) 0.017 & 0.868 \(\pm\) 0.016 \\
            & wine & 0.896 \(\pm\) 0.047 & 0.913 \(\pm\) 0.041 & \bf 0.932 \(\pm\) 0.038 & 0.930 \(\pm\) 0.040 & 0.912 \(\pm\) 0.054 & 0.916 \(\pm\) 0.051 & 0.917 \(\pm\) 0.051 & 0.912 \(\pm\) 0.050 & 0.922 \(\pm\) 0.048 & 0.924 \(\pm\) 0.049 \\
            & yeast & 0.911 \(\pm\) 0.010 & 0.946 \(\pm\) 0.005 & \bf 0.954 \(\pm\) 0.005 & 0.942 \(\pm\) 0.005 & 0.945 \(\pm\) 0.006 & 0.948 \(\pm\) 0.006 & 0.945 \(\pm\) 0.006 & 0.931 \(\pm\) 0.011 & 0.936 \(\pm\) 0.011 & 0.926 \(\pm\) 0.011 \\
            \midrule
            \multirow{18}{*}{\gls{plr}} & algae & 0.449 \(\pm\) 0.047 & 0.494 \(\pm\) 0.053 & 0.479 \(\pm\) 0.048 & 0.485 \(\pm\) 0.049 & \bf 0.497 \(\pm\) 0.057 & 0.474 \(\pm\) 0.047 & 0.479 \(\pm\) 0.050 & 0.494 \(\pm\) 0.047 & 0.484 \(\pm\) 0.046 & 0.488 \(\pm\) 0.050 \\
            & authorship & \bf 0.843 \(\pm\) 0.019 & 0.828 \(\pm\) 0.021 & 0.800 \(\pm\) 0.022 & 0.823 \(\pm\) 0.022 & 0.835 \(\pm\) 0.019 & 0.800 \(\pm\) 0.022 & 0.823 \(\pm\) 0.022 & 0.833 \(\pm\) 0.021 & 0.793 \(\pm\) 0.019 & 0.811 \(\pm\) 0.020 \\
            & blocks & \bf 0.956 \(\pm\) 0.004 & 0.948 \(\pm\) 0.005 & 0.846 \(\pm\) 0.008 & 0.927 \(\pm\) 0.006 & 0.955 \(\pm\) 0.005 & 0.842 \(\pm\) 0.008 & 0.927 \(\pm\) 0.006 & 0.953 \(\pm\) 0.005 & 0.847 \(\pm\) 0.008 & 0.930 \(\pm\) 0.006 \\
            & breast & 0.821 \(\pm\) 0.057 & 0.799 \(\pm\) 0.051 & 0.727 \(\pm\) 0.035 & 0.795 \(\pm\) 0.040 & \bf 0.822 \(\pm\) 0.047 & 0.732 \(\pm\) 0.040 & 0.795 \(\pm\) 0.040 & 0.801 \(\pm\) 0.058 & 0.726 \(\pm\) 0.040 & 0.790 \(\pm\) 0.047 \\
            & ecoli & \bf 0.815 \(\pm\) 0.025 & 0.774 \(\pm\) 0.025 & 0.682 \(\pm\) 0.028 & 0.767 \(\pm\) 0.028 & 0.798 \(\pm\) 0.026 & 0.669 \(\pm\) 0.027 & 0.745 \(\pm\) 0.029 & 0.785 \(\pm\) 0.026 & 0.677 \(\pm\) 0.030 & 0.751 \(\pm\) 0.026 \\
            & glass & \bf 0.812 \(\pm\) 0.030 & 0.802 \(\pm\) 0.034 & 0.765 \(\pm\) 0.027 & 0.811 \(\pm\) 0.031 & 0.803 \(\pm\) 0.031 & 0.761 \(\pm\) 0.031 & 0.807 \(\pm\) 0.034 & 0.806 \(\pm\) 0.028 & 0.773 \(\pm\) 0.030 & 0.809 \(\pm\) 0.033 \\
            & iris & \bf 0.932 \(\pm\) 0.031 & 0.930 \(\pm\) 0.035 & 0.855 \(\pm\) 0.049 & 0.897 \(\pm\) 0.044 & 0.925 \(\pm\) 0.036 & 0.855 \(\pm\) 0.047 & 0.896 \(\pm\) 0.045 & 0.930 \(\pm\) 0.032 & 0.867 \(\pm\) 0.051 & 0.900 \(\pm\) 0.044 \\
            & letter & 0.736 \(\pm\) 0.003 & 0.669 \(\pm\) 0.003 & 0.626 \(\pm\) 0.003 & 0.731 \(\pm\) 0.003 & 0.720 \(\pm\) 0.003 & 0.595 \(\pm\) 0.005 & 0.641 \(\pm\) 0.007 & \bf 0.737 \(\pm\) 0.003 & 0.642 \(\pm\) 0.004 & 0.730 \(\pm\) 0.003 \\
            & libras & 0.705 \(\pm\) 0.021 & 0.711 \(\pm\) 0.016 & 0.705 \(\pm\) 0.018 & \bf 0.724 \(\pm\) 0.019 & 0.705 \(\pm\) 0.019 & 0.703 \(\pm\) 0.018 & 0.721 \(\pm\) 0.017 & 0.712 \(\pm\) 0.020 & 0.695 \(\pm\) 0.018 & 0.711 \(\pm\) 0.019 \\
            & movies & \bf 0.427 \(\pm\) 0.034 & 0.358 \(\pm\) 0.030 & 0.308 \(\pm\) 0.032 & 0.244 \(\pm\) 0.030 & 0.369 \(\pm\) 0.036 & 0.333 \(\pm\) 0.038 & 0.269 \(\pm\) 0.037 & 0.366 \(\pm\) 0.029 & 0.336 \(\pm\) 0.036 & 0.249 \(\pm\) 0.030 \\
            & pendigits & \bf 0.847 \(\pm\) 0.005 & 0.779 \(\pm\) 0.005 & 0.603 \(\pm\) 0.004 & 0.776 \(\pm\) 0.005 & 0.834 \(\pm\) 0.005 & 0.581 \(\pm\) 0.006 & 0.708 \(\pm\) 0.007 & 0.817 \(\pm\) 0.005 & 0.635 \(\pm\) 0.005 & 0.785 \(\pm\) 0.006 \\
            & political & 0.695 \(\pm\) 0.019 & 0.671 \(\pm\) 0.016 & 0.643 \(\pm\) 0.020 & 0.671 \(\pm\) 0.019 & \bf 0.698 \(\pm\) 0.015 & 0.633 \(\pm\) 0.023 & 0.659 \(\pm\) 0.021 & 0.665 \(\pm\) 0.017 & 0.630 \(\pm\) 0.019 & 0.636 \(\pm\) 0.020 \\
            & satimage & \bf 0.896 \(\pm\) 0.005 & 0.871 \(\pm\) 0.005 & 0.753 \(\pm\) 0.007 & 0.855 \(\pm\) 0.005 & 0.885 \(\pm\) 0.004 & 0.743 \(\pm\) 0.009 & 0.836 \(\pm\) 0.006 & 0.879 \(\pm\) 0.005 & 0.753 \(\pm\) 0.006 & 0.852 \(\pm\) 0.006 \\
            & segment & \bf 0.918 \(\pm\) 0.010 & 0.894 \(\pm\) 0.009 & 0.734 \(\pm\) 0.012 & 0.876 \(\pm\) 0.010 & 0.911 \(\pm\) 0.011 & 0.723 \(\pm\) 0.010 & 0.848 \(\pm\) 0.012 & 0.910 \(\pm\) 0.008 & 0.773 \(\pm\) 0.013 & 0.881 \(\pm\) 0.010 \\
            & vehicle & \bf 0.820 \(\pm\) 0.018 & 0.812 \(\pm\) 0.020 & 0.779 \(\pm\) 0.020 & 0.802 \(\pm\) 0.023 & 0.807 \(\pm\) 0.020 & 0.766 \(\pm\) 0.022 & 0.792 \(\pm\) 0.022 & 0.816 \(\pm\) 0.019 & 0.770 \(\pm\) 0.024 & 0.792 \(\pm\) 0.022 \\
            & vowel & \bf 0.781 \(\pm\) 0.015 & 0.765 \(\pm\) 0.015 & 0.742 \(\pm\) 0.014 & 0.772 \(\pm\) 0.016 & 0.765 \(\pm\) 0.015 & 0.733 \(\pm\) 0.015 & 0.763 \(\pm\) 0.014 & 0.772 \(\pm\) 0.016 & 0.742 \(\pm\) 0.015 & 0.769 \(\pm\) 0.015 \\
            & wine & \bf 0.879 \(\pm\) 0.045 & 0.870 \(\pm\) 0.051 & 0.868 \(\pm\) 0.045 & 0.869 \(\pm\) 0.046 & 0.870 \(\pm\) 0.053 & 0.865 \(\pm\) 0.046 & 0.866 \(\pm\) 0.046 & 0.878 \(\pm\) 0.046 & 0.864 \(\pm\) 0.048 & 0.872 \(\pm\) 0.045 \\
            & yeast & \bf 0.817 \(\pm\) 0.009 & 0.807 \(\pm\) 0.010 & 0.772 \(\pm\) 0.007 & 0.807 \(\pm\) 0.009 & 0.812 \(\pm\) 0.009 & 0.771 \(\pm\) 0.007 & 0.804 \(\pm\) 0.009 & 0.808 \(\pm\) 0.007 & 0.764 \(\pm\) 0.007 & 0.802 \(\pm\) 0.007 \\
            \bottomrule
        \end{tabu}
    \label{table:accuracy_0}
    }
\end{table*}

\begin{table*}[!ht]
    \centering
    \caption{Accuracy results for incomplete rankings with 30\% of missing labels}
    \resizebox{\textwidth}{!}
    {
        \begin{tabu}{lrrrrrrrrrrr}
            \toprule
            Problem & Dataset & \gls{rpc} & \gls{st}-\gls{rr} & \gls{st}-\gls{pi} & \gls{st}-\gls{eps} & Chain-\gls{rr} & Chain-\gls{pi} & Chain-\gls{eps} & Native-\gls{rr} & Native-\gls{pi} & Native-\gls{eps} \\
            \midrule
            \multirow{13}{*}{\gls{lr}} & authorship & 0.866 \(\pm\) 0.025 & 0.909 \(\pm\) 0.021 & \bf 0.919 \(\pm\) 0.020 & 0.917 \(\pm\) 0.019 & 0.726 \(\pm\) 0.029 & 0.807 \(\pm\) 0.025 & 0.851 \(\pm\) 0.025 & 0.728 \(\pm\) 0.026 & 0.817 \(\pm\) 0.021 & 0.854 \(\pm\) 0.023 \\
            & glass & 0.823 \(\pm\) 0.053 & 0.885 \(\pm\) 0.035 & \bf 0.893 \(\pm\) 0.039 & 0.891 \(\pm\) 0.038 & 0.602 \(\pm\) 0.060 & 0.617 \(\pm\) 0.060 & 0.643 \(\pm\) 0.061 & 0.625 \(\pm\) 0.056 & 0.635 \(\pm\) 0.059 & 0.668 \(\pm\) 0.060 \\
            & iris & 0.927 \(\pm\) 0.042 & 0.955 \(\pm\) 0.042 & \bf 0.964 \(\pm\) 0.038 & 0.963 \(\pm\) 0.038 & 0.511 \(\pm\) 0.113 & 0.589 \(\pm\) 0.112 & 0.599 \(\pm\) 0.117 & 0.517 \(\pm\) 0.099 & 0.591 \(\pm\) 0.115 & 0.616 \(\pm\) 0.110 \\
            & letter & 0.879 \(\pm\) 0.002 & 0.934 \(\pm\) 0.001 & \bf 0.936 \(\pm\) 0.001 & 0.734 \(\pm\) 0.005 & 0.802 \(\pm\) 0.002 & 0.650 \(\pm\) 0.004 & 0.718 \(\pm\) 0.002 & 0.722 \(\pm\) 0.003 & 0.611 \(\pm\) 0.003 & 0.629 \(\pm\) 0.003 \\
            & libras & 0.784 \(\pm\) 0.022 & 0.878 \(\pm\) 0.014 & \bf 0.885 \(\pm\) 0.015 & 0.855 \(\pm\) 0.014 & 0.690 \(\pm\) 0.022 & 0.637 \(\pm\) 0.024 & 0.670 \(\pm\) 0.021 & 0.690 \(\pm\) 0.020 & 0.634 \(\pm\) 0.025 & 0.671 \(\pm\) 0.020 \\
            & movies & 0.206 \(\pm\) 0.031 & \bf 0.353 \(\pm\) 0.037 & 0.345 \(\pm\) 0.039 & 0.346 \(\pm\) 0.038 & 0.308 \(\pm\) 0.040 & 0.257 \(\pm\) 0.039 & 0.303 \(\pm\) 0.041 & 0.285 \(\pm\) 0.034 & 0.224 \(\pm\) 0.035 & 0.280 \(\pm\) 0.033 \\
            & pendigits & 0.949 \(\pm\) 0.001 & 0.968 \(\pm\) 0.001 & \bf 0.974 \(\pm\) 0.001 & 0.964 \(\pm\) 0.001 & 0.813 \(\pm\) 0.003 & 0.791 \(\pm\) 0.003 & 0.825 \(\pm\) 0.003 & 0.805 \(\pm\) 0.003 & 0.781 \(\pm\) 0.003 & 0.819 \(\pm\) 0.003 \\
            & political & 0.484 \(\pm\) 0.053 & 0.670 \(\pm\) 0.055 & 0.700 \(\pm\) 0.057 & \bf 0.700 \(\pm\) 0.058 & 0.505 \(\pm\) 0.063 & 0.505 \(\pm\) 0.059 & 0.546 \(\pm\) 0.062 & 0.506 \(\pm\) 0.059 & 0.493 \(\pm\) 0.070 & 0.556 \(\pm\) 0.062 \\
            & segment & 0.949 \(\pm\) 0.005 & 0.967 \(\pm\) 0.004 & \bf 0.972 \(\pm\) 0.004 & 0.969 \(\pm\) 0.004 & 0.714 \(\pm\) 0.011 & 0.726 \(\pm\) 0.012 & 0.753 \(\pm\) 0.012 & 0.713 \(\pm\) 0.012 & 0.728 \(\pm\) 0.013 & 0.755 \(\pm\) 0.013 \\
            & vehicle & 0.814 \(\pm\) 0.021 & 0.853 \(\pm\) 0.023 & 0.875 \(\pm\) 0.021 & \bf 0.877 \(\pm\) 0.021 & 0.603 \(\pm\) 0.039 & 0.658 \(\pm\) 0.034 & 0.686 \(\pm\) 0.035 & 0.621 \(\pm\) 0.036 & 0.680 \(\pm\) 0.030 & 0.714 \(\pm\) 0.037 \\
            & vowel & 0.719 \(\pm\) 0.027 & 0.853 \(\pm\) 0.014 & \bf 0.864 \(\pm\) 0.015 & 0.850 \(\pm\) 0.015 & 0.646 \(\pm\) 0.020 & 0.618 \(\pm\) 0.020 & 0.646 \(\pm\) 0.018 & 0.638 \(\pm\) 0.023 & 0.609 \(\pm\) 0.022 & 0.638 \(\pm\) 0.020 \\
            & wine & 0.842 \(\pm\) 0.066 & 0.898 \(\pm\) 0.050 & 0.925 \(\pm\) 0.046 & \bf 0.928 \(\pm\) 0.045 & 0.517 \(\pm\) 0.099 & 0.619 \(\pm\) 0.100 & 0.648 \(\pm\) 0.109 & 0.506 \(\pm\) 0.111 & 0.638 \(\pm\) 0.106 & 0.668 \(\pm\) 0.124 \\
            & yeast & 0.889 \(\pm\) 0.012 & 0.940 \(\pm\) 0.006 & \bf 0.949 \(\pm\) 0.005 & 0.936 \(\pm\) 0.005 & 0.762 \(\pm\) 0.013 & 0.739 \(\pm\) 0.013 & 0.771 \(\pm\) 0.013 & 0.740 \(\pm\) 0.013 & 0.719 \(\pm\) 0.014 & 0.751 \(\pm\) 0.014 \\
            \midrule
            \multirow{18}{*}{\gls{plr}} & algae & 0.407 \(\pm\) 0.048 & \bf 0.474 \(\pm\) 0.044 & 0.461 \(\pm\) 0.048 & 0.466 \(\pm\) 0.046 & 0.321 \(\pm\) 0.044 & 0.376 \(\pm\) 0.053 & 0.355 \(\pm\) 0.049 & 0.338 \(\pm\) 0.050 & 0.380 \(\pm\) 0.045 & 0.362 \(\pm\) 0.047 \\
            & authorship & \bf 0.828 \(\pm\) 0.018 & 0.824 \(\pm\) 0.021 & 0.797 \(\pm\) 0.022 & 0.818 \(\pm\) 0.023 & 0.659 \(\pm\) 0.029 & 0.772 \(\pm\) 0.021 & 0.747 \(\pm\) 0.022 & 0.656 \(\pm\) 0.031 & 0.763 \(\pm\) 0.024 & 0.733 \(\pm\) 0.021 \\
            & blocks & \bf 0.951 \(\pm\) 0.005 & 0.942 \(\pm\) 0.006 & 0.840 \(\pm\) 0.009 & 0.923 \(\pm\) 0.006 & 0.574 \(\pm\) 0.019 & 0.661 \(\pm\) 0.008 & 0.525 \(\pm\) 0.007 & 0.610 \(\pm\) 0.012 & 0.671 \(\pm\) 0.009 & 0.527 \(\pm\) 0.006 \\
            & breast & \bf 0.791 \(\pm\) 0.052 & 0.780 \(\pm\) 0.054 & 0.722 \(\pm\) 0.044 & 0.784 \(\pm\) 0.043 & 0.481 \(\pm\) 0.069 & 0.568 \(\pm\) 0.054 & 0.560 \(\pm\) 0.055 & 0.503 \(\pm\) 0.076 & 0.577 \(\pm\) 0.062 & 0.559 \(\pm\) 0.059 \\
            & ecoli & \bf 0.789 \(\pm\) 0.027 & 0.742 \(\pm\) 0.023 & 0.672 \(\pm\) 0.029 & 0.756 \(\pm\) 0.025 & 0.468 \(\pm\) 0.030 & 0.600 \(\pm\) 0.023 & 0.554 \(\pm\) 0.024 & 0.520 \(\pm\) 0.031 & 0.606 \(\pm\) 0.026 & 0.560 \(\pm\) 0.027 \\
            & glass & 0.769 \(\pm\) 0.045 & 0.784 \(\pm\) 0.036 & 0.754 \(\pm\) 0.029 & \bf 0.791 \(\pm\) 0.033 & 0.501 \(\pm\) 0.055 & 0.612 \(\pm\) 0.044 & 0.596 \(\pm\) 0.043 & 0.515 \(\pm\) 0.053 & 0.628 \(\pm\) 0.044 & 0.615 \(\pm\) 0.047 \\
            & iris & 0.915 \(\pm\) 0.041 & \bf 0.924 \(\pm\) 0.037 & 0.831 \(\pm\) 0.056 & 0.889 \(\pm\) 0.050 & 0.456 \(\pm\) 0.097 & 0.576 \(\pm\) 0.092 & 0.555 \(\pm\) 0.096 & 0.466 \(\pm\) 0.101 & 0.596 \(\pm\) 0.079 & 0.583 \(\pm\) 0.088 \\
            & letter & 0.716 \(\pm\) 0.004 & 0.625 \(\pm\) 0.004 & 0.613 \(\pm\) 0.004 & \bf 0.721 \(\pm\) 0.003 & 0.378 \(\pm\) 0.006 & 0.618 \(\pm\) 0.005 & 0.565 \(\pm\) 0.004 & 0.440 \(\pm\) 0.003 & 0.536 \(\pm\) 0.004 & 0.509 \(\pm\) 0.003 \\
            & libras & 0.645 \(\pm\) 0.021 & 0.682 \(\pm\) 0.018 & 0.685 \(\pm\) 0.018 & \bf 0.703 \(\pm\) 0.018 & 0.466 \(\pm\) 0.025 & 0.560 \(\pm\) 0.025 & 0.562 \(\pm\) 0.022 & 0.500 \(\pm\) 0.026 & 0.555 \(\pm\) 0.024 & 0.560 \(\pm\) 0.022 \\
            & movies & \bf 0.438 \(\pm\) 0.030 & 0.342 \(\pm\) 0.037 & 0.292 \(\pm\) 0.031 & 0.238 \(\pm\) 0.032 & 0.269 \(\pm\) 0.036 & 0.351 \(\pm\) 0.036 & 0.174 \(\pm\) 0.031 & 0.281 \(\pm\) 0.028 & 0.326 \(\pm\) 0.037 & 0.148 \(\pm\) 0.028 \\
            & pendigits & \bf 0.839 \(\pm\) 0.005 & 0.762 \(\pm\) 0.005 & 0.595 \(\pm\) 0.005 & 0.769 \(\pm\) 0.006 & 0.497 \(\pm\) 0.008 & 0.698 \(\pm\) 0.007 & 0.521 \(\pm\) 0.005 & 0.586 \(\pm\) 0.005 & 0.708 \(\pm\) 0.006 & 0.530 \(\pm\) 0.004 \\
            & political & \bf 0.675 \(\pm\) 0.019 & 0.662 \(\pm\) 0.019 & 0.636 \(\pm\) 0.020 & 0.662 \(\pm\) 0.020 & 0.522 \(\pm\) 0.025 & 0.615 \(\pm\) 0.022 & 0.556 \(\pm\) 0.022 & 0.540 \(\pm\) 0.021 & 0.597 \(\pm\) 0.021 & 0.542 \(\pm\) 0.021 \\
            & satimage & \bf 0.887 \(\pm\) 0.005 & 0.862 \(\pm\) 0.005 & 0.746 \(\pm\) 0.007 & 0.850 \(\pm\) 0.005 & 0.615 \(\pm\) 0.011 & 0.742 \(\pm\) 0.006 & 0.652 \(\pm\) 0.007 & 0.691 \(\pm\) 0.008 & 0.762 \(\pm\) 0.006 & 0.658 \(\pm\) 0.006 \\
            & segment & \bf 0.907 \(\pm\) 0.011 & 0.871 \(\pm\) 0.011 & 0.718 \(\pm\) 0.011 & 0.865 \(\pm\) 0.012 & 0.506 \(\pm\) 0.018 & 0.647 \(\pm\) 0.011 & 0.530 \(\pm\) 0.010 & 0.576 \(\pm\) 0.014 & 0.658 \(\pm\) 0.011 & 0.539 \(\pm\) 0.010 \\
            & vehicle & 0.792 \(\pm\) 0.020 & \bf 0.799 \(\pm\) 0.022 & 0.770 \(\pm\) 0.022 & 0.791 \(\pm\) 0.020 & 0.538 \(\pm\) 0.033 & 0.635 \(\pm\) 0.026 & 0.626 \(\pm\) 0.028 & 0.543 \(\pm\) 0.034 & 0.644 \(\pm\) 0.028 & 0.623 \(\pm\) 0.031 \\
            & vowel & 0.745 \(\pm\) 0.014 & 0.741 \(\pm\) 0.014 & 0.728 \(\pm\) 0.014 & \bf 0.755 \(\pm\) 0.015 & 0.487 \(\pm\) 0.024 & 0.615 \(\pm\) 0.019 & 0.604 \(\pm\) 0.018 & 0.513 \(\pm\) 0.025 & 0.605 \(\pm\) 0.020 & 0.596 \(\pm\) 0.020 \\
            & wine & 0.847 \(\pm\) 0.036 & 0.855 \(\pm\) 0.039 & 0.860 \(\pm\) 0.042 & \bf 0.864 \(\pm\) 0.040 & 0.539 \(\pm\) 0.113 & 0.660 \(\pm\) 0.082 & 0.654 \(\pm\) 0.095 & 0.514 \(\pm\) 0.096 & 0.668 \(\pm\) 0.074 & 0.662 \(\pm\) 0.082 \\
            & yeast & \bf 0.801 \(\pm\) 0.011 & 0.795 \(\pm\) 0.009 & 0.767 \(\pm\) 0.006 & 0.799 \(\pm\) 0.007 & 0.542 \(\pm\) 0.021 & 0.700 \(\pm\) 0.010 & 0.681 \(\pm\) 0.010 & 0.614 \(\pm\) 0.013 & 0.694 \(\pm\) 0.013 & 0.675 \(\pm\) 0.012 \\
            \bottomrule
        \end{tabu}
    \label{table:accuracy_30}
    }
\end{table*}

\begin{table*}[!ht]
    \centering
    \caption{Accuracy results for incomplete rankings with 60\% of missing labels}
    \resizebox{\textwidth}{!}
    {
        \begin{tabu}{lrrrrrrrrrrr}
            \toprule
            Problem & Dataset & \gls{rpc} & \gls{st}-\gls{rr} & \gls{st}-\gls{pi} & \gls{st}-\gls{eps} & Chain-\gls{rr} & Chain-\gls{pi} & Chain-\gls{eps} & Native-\gls{rr} & Native-\gls{pi} & Native-\gls{eps} \\
            \midrule
            \multirow{13}{*}{\gls{lr}} & authorship & 0.814 \(\pm\) 0.028 & 0.893 \(\pm\) 0.021 & \bf 0.909 \(\pm\) 0.020 & 0.909 \(\pm\) 0.019 & 0.567 \(\pm\) 0.034 & 0.629 \(\pm\) 0.037 & 0.722 \(\pm\) 0.033 & 0.609 \(\pm\) 0.026 & 0.667 \(\pm\) 0.034 & 0.777 \(\pm\) 0.029 \\
            & glass & 0.743 \(\pm\) 0.058 & 0.851 \(\pm\) 0.042 & \bf 0.863 \(\pm\) 0.040 & 0.861 \(\pm\) 0.043 & 0.394 \(\pm\) 0.068 & 0.389 \(\pm\) 0.073 & 0.436 \(\pm\) 0.078 & 0.424 \(\pm\) 0.065 & 0.416 \(\pm\) 0.069 & 0.475 \(\pm\) 0.068 \\
            & iris & 0.860 \(\pm\) 0.056 & 0.938 \(\pm\) 0.046 & \bf 0.947 \(\pm\) 0.048 & 0.946 \(\pm\) 0.049 & 0.326 \(\pm\) 0.139 & 0.341 \(\pm\) 0.134 & 0.359 \(\pm\) 0.139 & 0.321 \(\pm\) 0.142 & 0.325 \(\pm\) 0.135 & 0.335 \(\pm\) 0.148 \\
            & letter & 0.834 \(\pm\) 0.002 & 0.923 \(\pm\) 0.001 & \bf 0.924 \(\pm\) 0.001 & 0.761 \(\pm\) 0.004 & 0.687 \(\pm\) 0.003 & 0.293 \(\pm\) 0.006 & 0.573 \(\pm\) 0.006 & 0.579 \(\pm\) 0.003 & 0.305 \(\pm\) 0.004 & 0.462 \(\pm\) 0.003 \\
            & libras & 0.632 \(\pm\) 0.027 & 0.846 \(\pm\) 0.016 & \bf 0.850 \(\pm\) 0.016 & 0.824 \(\pm\) 0.015 & 0.504 \(\pm\) 0.029 & 0.358 \(\pm\) 0.027 & 0.478 \(\pm\) 0.029 & 0.529 \(\pm\) 0.026 & 0.335 \(\pm\) 0.035 & 0.505 \(\pm\) 0.027 \\
            & movies & 0.204 \(\pm\) 0.035 & \bf 0.344 \(\pm\) 0.032 & 0.337 \(\pm\) 0.030 & 0.339 \(\pm\) 0.029 & 0.165 \(\pm\) 0.040 & 0.107 \(\pm\) 0.030 & 0.161 \(\pm\) 0.041 & 0.179 \(\pm\) 0.035 & 0.122 \(\pm\) 0.028 & 0.177 \(\pm\) 0.036 \\
            & pendigits & 0.927 \(\pm\) 0.002 & 0.962 \(\pm\) 0.001 & \bf 0.970 \(\pm\) 0.001 & 0.958 \(\pm\) 0.001 & 0.652 \(\pm\) 0.005 & 0.567 \(\pm\) 0.007 & 0.687 \(\pm\) 0.005 & 0.664 \(\pm\) 0.005 & 0.571 \(\pm\) 0.007 & 0.705 \(\pm\) 0.005 \\
            & political & 0.393 \(\pm\) 0.061 & 0.643 \(\pm\) 0.059 & 0.673 \(\pm\) 0.059 & \bf 0.678 \(\pm\) 0.059 & 0.354 \(\pm\) 0.063 & 0.332 \(\pm\) 0.069 & 0.385 \(\pm\) 0.070 & 0.408 \(\pm\) 0.060 & 0.347 \(\pm\) 0.053 & 0.447 \(\pm\) 0.063 \\
            & segment & 0.925 \(\pm\) 0.006 & 0.958 \(\pm\) 0.005 & \bf 0.964 \(\pm\) 0.005 & 0.961 \(\pm\) 0.004 & 0.516 \(\pm\) 0.016 & 0.500 \(\pm\) 0.016 & 0.565 \(\pm\) 0.016 & 0.525 \(\pm\) 0.016 & 0.506 \(\pm\) 0.018 & 0.584 \(\pm\) 0.018 \\
            & vehicle & 0.766 \(\pm\) 0.030 & 0.836 \(\pm\) 0.026 & \bf 0.863 \(\pm\) 0.024 & 0.862 \(\pm\) 0.024 & 0.457 \(\pm\) 0.048 & 0.486 \(\pm\) 0.051 & 0.533 \(\pm\) 0.056 & 0.478 \(\pm\) 0.043 & 0.518 \(\pm\) 0.044 & 0.569 \(\pm\) 0.047 \\
            & vowel & 0.494 \(\pm\) 0.032 & 0.810 \(\pm\) 0.018 & \bf 0.822 \(\pm\) 0.018 & 0.808 \(\pm\) 0.017 & 0.437 \(\pm\) 0.027 & 0.370 \(\pm\) 0.030 & 0.440 \(\pm\) 0.026 & 0.440 \(\pm\) 0.026 & 0.372 \(\pm\) 0.025 & 0.447 \(\pm\) 0.026 \\
            & wine & 0.715 \(\pm\) 0.113 & 0.884 \(\pm\) 0.061 & 0.912 \(\pm\) 0.059 & \bf 0.915 \(\pm\) 0.056 & 0.396 \(\pm\) 0.114 & 0.410 \(\pm\) 0.130 & 0.433 \(\pm\) 0.142 & 0.434 \(\pm\) 0.102 & 0.469 \(\pm\) 0.126 & 0.497 \(\pm\) 0.145 \\
            & yeast & 0.846 \(\pm\) 0.014 & 0.929 \(\pm\) 0.007 & \bf 0.939 \(\pm\) 0.007 & 0.924 \(\pm\) 0.006 & 0.592 \(\pm\) 0.016 & 0.515 \(\pm\) 0.022 & 0.618 \(\pm\) 0.017 & 0.592 \(\pm\) 0.014 & 0.509 \(\pm\) 0.019 & 0.617 \(\pm\) 0.014 \\
            \midrule
            \multirow{18}{*}{\gls{plr}} & algae & 0.342 \(\pm\) 0.046 & \bf 0.444 \(\pm\) 0.048 & 0.431 \(\pm\) 0.044 & 0.436 \(\pm\) 0.045 & 0.230 \(\pm\) 0.048 & 0.286 \(\pm\) 0.045 & 0.259 \(\pm\) 0.051 & 0.253 \(\pm\) 0.048 & 0.297 \(\pm\) 0.046 & 0.281 \(\pm\) 0.053 \\
            & authorship & 0.778 \(\pm\) 0.026 & \bf 0.806 \(\pm\) 0.018 & 0.787 \(\pm\) 0.022 & 0.803 \(\pm\) 0.022 & 0.509 \(\pm\) 0.041 & 0.645 \(\pm\) 0.041 & 0.648 \(\pm\) 0.033 & 0.542 \(\pm\) 0.031 & 0.679 \(\pm\) 0.027 & 0.682 \(\pm\) 0.024 \\
            & blocks & \bf 0.943 \(\pm\) 0.005 & 0.933 \(\pm\) 0.006 & 0.830 \(\pm\) 0.010 & 0.917 \(\pm\) 0.007 & 0.418 \(\pm\) 0.012 & 0.619 \(\pm\) 0.018 & 0.453 \(\pm\) 0.010 & 0.434 \(\pm\) 0.008 & 0.610 \(\pm\) 0.010 & 0.455 \(\pm\) 0.008 \\
            & breast & 0.671 \(\pm\) 0.065 & 0.754 \(\pm\) 0.054 & 0.715 \(\pm\) 0.042 & \bf 0.763 \(\pm\) 0.047 & 0.297 \(\pm\) 0.121 & 0.378 \(\pm\) 0.100 & 0.362 \(\pm\) 0.106 & 0.310 \(\pm\) 0.122 & 0.405 \(\pm\) 0.107 & 0.391 \(\pm\) 0.121 \\
            & ecoli & \bf 0.739 \(\pm\) 0.038 & 0.713 \(\pm\) 0.025 & 0.652 \(\pm\) 0.027 & 0.731 \(\pm\) 0.025 & 0.372 \(\pm\) 0.035 & 0.498 \(\pm\) 0.036 & 0.443 \(\pm\) 0.029 & 0.406 \(\pm\) 0.037 & 0.522 \(\pm\) 0.037 & 0.461 \(\pm\) 0.032 \\
            & glass & 0.666 \(\pm\) 0.054 & 0.756 \(\pm\) 0.039 & 0.726 \(\pm\) 0.039 & \bf 0.759 \(\pm\) 0.043 & 0.344 \(\pm\) 0.061 & 0.447 \(\pm\) 0.055 & 0.435 \(\pm\) 0.062 & 0.391 \(\pm\) 0.052 & 0.473 \(\pm\) 0.053 & 0.466 \(\pm\) 0.049 \\
            & iris & 0.873 \(\pm\) 0.054 & \bf 0.893 \(\pm\) 0.046 & 0.816 \(\pm\) 0.043 & 0.878 \(\pm\) 0.043 & 0.297 \(\pm\) 0.129 & 0.345 \(\pm\) 0.135 & 0.333 \(\pm\) 0.133 & 0.268 \(\pm\) 0.119 & 0.360 \(\pm\) 0.142 & 0.331 \(\pm\) 0.146 \\
            & letter & 0.676 \(\pm\) 0.005 & 0.580 \(\pm\) 0.004 & 0.595 \(\pm\) 0.003 & \bf 0.705 \(\pm\) 0.003 & 0.322 \(\pm\) 0.007 & 0.535 \(\pm\) 0.006 & 0.511 \(\pm\) 0.004 & 0.351 \(\pm\) 0.002 & 0.435 \(\pm\) 0.006 & 0.398 \(\pm\) 0.003 \\
            & libras & 0.529 \(\pm\) 0.028 & 0.637 \(\pm\) 0.021 & 0.649 \(\pm\) 0.018 & \bf 0.665 \(\pm\) 0.019 & 0.297 \(\pm\) 0.030 & 0.364 \(\pm\) 0.028 & 0.398 \(\pm\) 0.025 & 0.331 \(\pm\) 0.027 & 0.363 \(\pm\) 0.028 & 0.424 \(\pm\) 0.026 \\
            & movies & \bf 0.444 \(\pm\) 0.031 & 0.335 \(\pm\) 0.030 & 0.272 \(\pm\) 0.029 & 0.226 \(\pm\) 0.027 & 0.244 \(\pm\) 0.034 & 0.317 \(\pm\) 0.039 & 0.143 \(\pm\) 0.030 & 0.276 \(\pm\) 0.035 & 0.310 \(\pm\) 0.036 & 0.134 \(\pm\) 0.034 \\
            & pendigits & \bf 0.822 \(\pm\) 0.005 & 0.738 \(\pm\) 0.004 & 0.584 \(\pm\) 0.005 & 0.757 \(\pm\) 0.006 & 0.443 \(\pm\) 0.006 & 0.668 \(\pm\) 0.006 & 0.438 \(\pm\) 0.005 & 0.471 \(\pm\) 0.005 & 0.676 \(\pm\) 0.007 & 0.443 \(\pm\) 0.005 \\
            & political & 0.629 \(\pm\) 0.023 & 0.644 \(\pm\) 0.019 & 0.626 \(\pm\) 0.021 & \bf 0.649 \(\pm\) 0.021 & 0.411 \(\pm\) 0.026 & 0.489 \(\pm\) 0.028 & 0.444 \(\pm\) 0.027 & 0.449 \(\pm\) 0.028 & 0.490 \(\pm\) 0.025 & 0.458 \(\pm\) 0.025 \\
            & satimage & \bf 0.869 \(\pm\) 0.006 & 0.848 \(\pm\) 0.005 & 0.733 \(\pm\) 0.007 & 0.840 \(\pm\) 0.006 & 0.494 \(\pm\) 0.011 & 0.666 \(\pm\) 0.009 & 0.572 \(\pm\) 0.008 & 0.523 \(\pm\) 0.009 & 0.701 \(\pm\) 0.008 & 0.593 \(\pm\) 0.007 \\
            & segment & \bf 0.889 \(\pm\) 0.011 & 0.847 \(\pm\) 0.011 & 0.691 \(\pm\) 0.013 & 0.849 \(\pm\) 0.012 & 0.407 \(\pm\) 0.015 & 0.567 \(\pm\) 0.017 & 0.431 \(\pm\) 0.011 & 0.426 \(\pm\) 0.013 & 0.588 \(\pm\) 0.016 & 0.443 \(\pm\) 0.010 \\
            & vehicle & 0.707 \(\pm\) 0.031 & \bf 0.773 \(\pm\) 0.025 & 0.751 \(\pm\) 0.022 & 0.770 \(\pm\) 0.023 & 0.376 \(\pm\) 0.039 & 0.449 \(\pm\) 0.047 & 0.441 \(\pm\) 0.057 & 0.404 \(\pm\) 0.044 & 0.483 \(\pm\) 0.045 & 0.479 \(\pm\) 0.044 \\
            & vowel & 0.632 \(\pm\) 0.025 & 0.705 \(\pm\) 0.016 & 0.704 \(\pm\) 0.016 & \bf 0.727 \(\pm\) 0.017 & 0.334 \(\pm\) 0.028 & 0.435 \(\pm\) 0.023 & 0.446 \(\pm\) 0.024 & 0.355 \(\pm\) 0.026 & 0.441 \(\pm\) 0.021 & 0.454 \(\pm\) 0.021 \\
            & wine & 0.747 \(\pm\) 0.064 & 0.817 \(\pm\) 0.052 & 0.855 \(\pm\) 0.049 & \bf 0.862 \(\pm\) 0.048 & 0.351 \(\pm\) 0.108 & 0.397 \(\pm\) 0.120 & 0.383 \(\pm\) 0.125 & 0.385 \(\pm\) 0.104 & 0.434 \(\pm\) 0.130 & 0.426 \(\pm\) 0.137 \\
            & yeast & 0.767 \(\pm\) 0.013 & 0.777 \(\pm\) 0.008 & 0.758 \(\pm\) 0.007 & \bf 0.788 \(\pm\) 0.008 & 0.449 \(\pm\) 0.016 & 0.557 \(\pm\) 0.021 & 0.574 \(\pm\) 0.016 & 0.489 \(\pm\) 0.015 & 0.568 \(\pm\) 0.015 & 0.582 \(\pm\) 0.012 \\
            \bottomrule
        \end{tabu}
    \label{table:accuracy_60}
    }
\end{table*}

\end{document}